\documentclass{article}

\title{\Large\textbf{On Adaptive Estimation for Dynamic Bernoulli Bandits }}

\author{
	\textbf{Xue Lu} \\ \textbf{xue.lu12@imperial.ac.uk}  \\ Department of Mathematics \\ Imperial College London \\ London, SW7 2AZ         \and
        \textbf{Niall Adams} \\ \textbf{n.adams@imperial.ac.uk} \\ Department of Mathematics \\ Imperial College London \\ London, SW7 2AZ  \and 
        \textbf{Nikolas Kantas} \\ \textbf{n.kantas@imperial.ac.uk} \\ Department of Mathematics \\ Imperial College London \\ London, SW7 2AZ 
        }
        
\date{}

\usepackage[top=2.5cm, bottom=3cm, left=2cm, right=2cm]{geometry}
\usepackage{amssymb,amsmath}
\usepackage{graphicx}
\usepackage{subfig}
\usepackage{caption}
\usepackage{upquote}
\usepackage{xcolor}
\usepackage{fancyvrb}
\usepackage{multicol}
\usepackage{listings}
\usepackage{upgreek}
\usepackage{multirow}
\usepackage{url} 
\usepackage{algorithm}
\usepackage{algorithmic}
\usepackage{booktabs}
\usepackage{enumitem}
\usepackage{adjustbox}
\usepackage{rotating}
\usepackage{pdflscape}
\usepackage[authoryear, round]{natbib}

\DeclareMathOperator*{\argmax}{arg\,max}

\begin{document}
\maketitle

\begin{abstract}
The multi-armed bandit (MAB) problem is a classic example of the exploration-exploitation
dilemma. It is concerned with maximising the total rewards for a gambler by sequentially
pulling an arm from a multi-armed slot machine where each arm is associated with
a reward distribution. In static MABs, the reward distributions
do not change over time, while in dynamic MABs, each arm's reward distribution
can change, and the optimal arm can switch over time. Motivated by many real applications where rewards are binary, we focus on dynamic Bernoulli bandits. Standard methods like $\epsilon$-Greedy and Upper Confidence Bound (UCB), which rely on the sample mean estimator, often fail to track changes in the underlying reward for dynamic problems. In this paper, we overcome the shortcoming of slow response to change by deploying adaptive estimation in the standard methods and propose a new family of algorithms, which are adaptive versions of $\epsilon$-Greedy, UCB, and Thompson sampling. These new methods are simple and easy to implement. Moreover, they do not require any prior knowledge about the dynamic reward process, which is important for real applications. We examine the new algorithms numerically in different scenarios and the results show solid improvements of our algorithms in dynamic environments.

\end{abstract}

\section{Introduction}
\label{sec-intro}
The multi-armed bandit (MAB) problem is a classic decision problem where one needs to balance acquiring new knowledge with optimising the
choices based on current knowledge, a dilemma commonly referred to as the exploration-exploitation trade-off. The problem originally proposed by \citet{Robbins1952} aims to sequentially make selections among a (finite) set of arms, $\mathcal{A}$, and maximise the total reward obtained through selections during a (possibly infinite) time horizon $T$.
The MAB framework is natural to model many real-world problems. It was originally motivated by the design of clinical trials (\citealp{Thompson1933}; see also \citealp{Press2009}, and \citealp{Villar2015}, for some recent developments). Other applications include online advertising \citep{Li2010, Scott2015}, adaptive routing \citep{Awerbuch2008}, and  financial portfolio design \citep{Brochu2011, Shen2015}. 
In stochastic MABs, each arm $a \in \mathcal{A}$ is characterised by an unknown reward distribution. The Bernoulli distribution is a very natural choice that appears very often in the literature, because in many real applications, the reward can be represented by 0 or 1.
For example,  in clinical trials, we obtain a reward 1 for a successful treatment, and a reward 0 otherwise \citep{Villar2015}; in online advertising, counts of clicks are often used to measure success \citep{Scott2010}. 

Formally, the MAB problem may be stated as follows: for discrete time $t = 1, \cdots, T$, the decision maker selects one arm $a_{t}$ from $\mathcal{A}$ and receives a reward $Y_{t}(a_{t})$.  The goal is to optimise the arm selection sequence and
maximise the total expected reward $\sum_{t=1}^{T}  \mathbb{E} \left [Y_{t}(a_{t}) \right ]$, or equivalently, minimise the total regret:
\begin{align}
\label{seq-regret}
R_{T} &=  \sum_{t=1}^{T} \mathbb{E} [Y_{t}(a_{t}^{*})] - \sum_{t=1}^{T} \mathbb{E} [Y_{t}(a_{t})], \\
\notag
a_{t}^{*} &= \argmax_{a' \in \mathcal{A}}\mathbb{E} \left [Y_{t}(a') \right ],
\end{align}
where $a_{t}^{*}$
is the optimal arm at time $t$. 
The total regret can be interpreted as the difference between the total expected reward obtained by playing an optimal strategy (selecting the optimal arm at every step) and that obtained by the algorithm.
For notational convenience, we let $\mu_{t}(a), a \in \mathcal{A}$, denote the expected reward of arm $a$ at time $t$, i.e., $\mu_{t}(a) = \mathbb{E} \left [Y_{t}(a) \right ]$. 
In the rest of this paper, we will also use notations like $Y_{t}$ and $\mu_{t}$ when we introduce the methods that can be applied separately to different arms.

The classic MAB problem assumes the reward distribution structure \textit{does not change} over time. That is to say, in this case, the optimal arm is the same for all $t$. A MAB problem with static reward distributions is also known as the \textit{stationary}, or \textit{static} MAB problem in the literature \citep[e.g.,][]{Garivier2011, Slivkins2008}. A dynamic MAB, where changes occur in the underlying reward distributions, is more realistic in real-world applications such as online advertising. An agent always seeks the best web position (that is, the placement of the advertisement on a webpage), and/or advertisement content, to maximise the probability of obtaining clicks. 
However, due to inherent changes in marketplace, the optimal choice may change over time, and thus the assumption of static reward distributions is not adequate in this example.

Two main types of change have been studied in the literature of dynamic MAB: \textit{abrupt changes} \citep{Garivier2011, Yu2009}, and \textit{drifting} \citep{Granmo2010, Gupta2011, Slivkins2008}. For abrupt changes, the expected reward of an arm remains constant for some period and changes suddenly at possibly unknown times \citep{Garivier2011}. The study of drifting dynamic bandits follows the seminal work of \citet{Whittle1988}, in which \textit{restless bandits} were introduced. In Whittle's study, the state of an arm can change according to a Markov transition function over time whether it is selected or not. Restless bandits are regarded as intractable, i.e., it is not possible to derive an optimal strategy even if the transitions are deterministic \citep{Papadimitriou1999}.  In recent studies of drifting dynamic bandits, the expected reward of an arm is often modelled by a random walk \citep[e.g.,][]{Granmo2010, Gupta2011, Slivkins2008}. 

In this work, we look at the problem of dynamic bandits where the expectation of the reward distribution changes over time, focusing on the Bernoulli reward distribution because of its wide relevance in real applications. In addition, we will emphasise cases where the changes of the reward distribution can \textit{really have} an effect on the decision making. As an example, for a two-armed Bernoulli bandit, the expected reward of Arm 1 oscillates in $[0.1, 0.3]$ over time, and the expected reward of Arm 2 oscillates in $[0.8, 0.9]$; the reward distributions for both arms change, but the optimal arm remains the same. We will not regard this example as a dynamic case. 

Many algorithms have been proposed in the literature to perform arm selection for MAB. Some of the most popular include $\epsilon$-Greedy \citep{Watkins1989}, Upper Confidence Bound \citep[UCB;][]{Auer2002c}, and Thompson Sampling \citep[TS;][]{Thompson1933}. These methods have been extended in various ways to improve performance. For example, \citet{Garivier2011a} proposed the Kullback-Leibler UCB (KL-UCB) method which satisfies a uniformly better regret bound than UCB. \citet{May2012a} introduced the Optimistic Thompson Sampling (OTS) method to boost exploration in TS. Extensions for dynamic bandits will be described in Section~\ref{sec-selection}. Even in their basic forms, $\epsilon$-Greedy, UCB and TS can perform well in practice in many situations \citep[e.g.,][]{Chapelle2011, Kuleshov2014, Vermorel2005}. One thing that these methods have in common is that, they treat all the observations $Y_{1}, \cdots, Y_{t}$ equally when estimating or making inference of $\mu_{t}$. Specifically, $\epsilon$-Greedy and UCB use sample averages to estimate $\mu_t$. In static cases, given that $Y_{1}, \cdots, Y_{t}$ are i.i.d, this choice is sensible from a theoretical perspective, and one could invoke various asymptotic results as justification (e.g., law of large numbers, central limit theorem, Berry Essen inequality etc.).  From a practical point of view, when $\mu_t$ changes significantly with time, it could become a bottleneck in performance. The problem is that a sample average does not put more weight on more recent data $Y_{t}$, which is a direct observation of $\mu_{t}$.
In this paper we will consider using a different estimator for $\mu_t$ that is inspired from adaptive estimation \citep{Haykin2002} and propose novel modifications of popular MAB algorithms.

\subsection{Contributions and Organisation}

We propose algorithms that use Adaptive Forgetting Factors \citep[AFFs;][]{Bodenham2016} in conjunction with the standard MAB methods. 
In addition to using AFFs in estimating the mean, there is extra information obtained from the computation that can be used in our modification of the exploration schemes.
This results in a new
family of algorithms for dynamic Bernoulli bandits. 
These algorithms overcome the shortcomings related to using sample averages to estimate dynamically changing rewards. Our methods are easy to implement and require very little tuning effort; they are quite robust to tuning parameters and their initialisation does not require assumptions or knowledge about the model structure in advance. Furthermore, the extent of exploration in our algorithms is adjusted according to the total number of arms, and the performance gains are substantial in the case that the number of arms is large.

The remainder of this paper is structured as follows: Section~\ref{sec-AdaEst} briefly summarises some adaptive estimation techniques, focusing on AFFs. Section~\ref{sec-selection} introduces the methodology for arm selection. Section~\ref{sec-exps} presents a variety of numerical results for different dynamic models and MAB algorithms. We summarise our findings in Section~\ref{sec-con}.

\section{Adaptive Estimation Using Forgetting Factors}
\label{sec-AdaEst}

Solving the MAB problem involves two main steps: learning the reward distribution of each arm (\textit{estimation step}), and selecting one arm to play (\textit{selection step}). The foundation of making a good selection is to correctly and efficiently track the expected reward of the arms, especially in the context of time-evolving reward distributions. Adaptive estimation approaches are useful for this task as they provide an estimator that follows a moving target \citep{Anagnostopoulos2012a, Bodenham2016}, here the target is the expected reward.
In this section, we introduce how to use an AFF estimator for monitoring a single arm. For the sake of simplicity, when clear we drop dependence on arms in the notation.


Assume now that we select one arm all the time until $t$ and receive rewards $Y_{1}, \cdots, Y_{t}$. If the reward distribution is static, $Y_{1}, \cdots, Y_{t}$ are i.i.d. Therefore, it is natural to estimate the expected reward via the sample mean: $\bar{Y}_{t} = \frac{1}{t} \sum_{i=1}^{t} Y_{i}$. This sample mean estimator was widely used in the algorithms designed for the static MAB problem such as $\epsilon$-Greedy and UCB. One problem with this estimator is that it often fails in the case that the reward distribution changes over time. The adaptive filtering literature \citep{Haykin2002} provides a generic and practical tool to track a time-evolving data stream, and it has been recently adapted to a variety of streaming machine learning problems \citep{Anagnostopoulos2012a, Bodenham2016}. The key idea behind adaptive estimation is to gradually reduce the weight on older data as new data arrives \citep{Haykin2002}. For example, a fixed forgetting factor estimator employs a forgetting/discounting factor $\lambda$, $\lambda \in [0, 1]$, and takes the form $\hat{Y}_{\lambda, t} = \frac{1}{w_{\lambda, t}} \sum_{i=1}^{t} \lambda^{t-i} Y_{i} $, where $w_{\lambda, t}$ is a normalising constant. \cite{Bodenham2016} illustrated that the fixed forgetting factor estimator has some similarities with the Exponentially Weighted Moving Average (EWMA) scheme \citep{Roberts1959} which is a basic approach in the change detection literature \citep{Tsung2010}. 

In this paper, we will use an adaptive forgetting factor where the magnitude of the forgetting factor $\lambda$ can be adjusted at each time step for better adaptation. One main advantage of an AFF estimator is that it can respond quickly to the changes of a target without requiring any prior knowledge about the process. In addition, by using data-adaptive tuning of $\lambda$, we side-step the problem of setting a key control parameter. Therefore, it is very useful when applied to dynamic MABs where we do not have any knowledge about the reward process. 

Our AFF formulation follows \cite{Bodenham2016}. We present only the main methodology. For the observed reward sequence (of a single arm) $Y_{1}, \cdots, Y_{t}$, the adaptive forgetting factor mean (denoted by $\hat{Y}_{t}$) is  defined as follows:
 \begin{align}
 \label{eqt-AFFmean}
\hat{Y}_{t} &= \frac{1}{w_{t}} \sum_{i=1}^{t} \left(\prod_{p=i}^{t-1} \lambda_{p} \right) Y_{i},
\end{align}
where the normalising constant $w_{t} = \sum_{i=1}^{t} \left(\prod_{p=i}^{t-1} \lambda_{p} \right)$ is selected to give unbiased estimation when $Y_{1}, \cdots, Y_{t}$ are i.i.d. For convenience, we set the empty product $\prod_{p=t}^{t-1} \lambda_{p} = 1$.
We can update $\hat{Y}_{t}$ via the following recursive updating equations:
\begin{align}
\label{eqt-hatyt}
\hat{Y}_{t}  &=  \frac{m_{t}}{w_{t}}, \\
\label{eqt-mt}
m_{t} & =  \lambda_{t-1} m_{t-1} + Y_{t} , \\
\label{eqt-wt}
w_{t} & =  \lambda_{t-1} w_{t-1} + 1.
\end{align}
The adaptive forgetting factor $\boldsymbol{\lambda}_{t} = (\lambda_{1}, \lambda_{2}, \cdots, \lambda_{t})$ is a expanding sequence over time, and the forgetting factor $\lambda_{t}$ is computed via a single gradient descent step, which is 
\begin{align}
\label{eqt-lambda}
\lambda_{t} = \lambda_{t-1} - \eta \Delta(L_{t}, {\boldsymbol{\lambda}_{t-2}}),
\end{align}
where $\eta$ ($\eta \ll 1$) is the step size, and $L_{t}$ is a user determined cost function of the estimator $\hat{Y}_{t}$. Here, we choose $L_{t} = (\hat{Y}_{t-1} - Y_{t})^{2}$ for good mean tracking performance, which can be interpreted as the one-step-ahead squared prediction error. Other choices are possible, such as the one-step-ahead negative log-likelihood \citep{Anagnostopoulos2012a}, but this will not be pursued here. In addition, $\Delta(L_{t}, {\boldsymbol{\lambda}_{t-2}})$ is a derivative-like function of $L_{t}$ with respect to $\boldsymbol{\lambda}_{t-2}$ \citep[see][sect.~4.2.1 for details]{Bodenham2016}. Note, the index of $\boldsymbol{\lambda}$ is $(t-2)$ as only $\lambda_{1}, \cdots, \lambda_{t-2}$ are involved in $L_{t}$; if the value $\lambda_{t}$ computed via (\ref{eqt-lambda}) is greater than 1 (or less than 0), we truncate it to 1 (or 0) to ensure that $\lambda_{t} \in [0,1]$. We require the following recursions to sequentially compute $\Delta(L_{t}, {\boldsymbol{\lambda}_{t-2}})$:
\begin{align}
\dot{m}_{t} &= \lambda_{t-1} \dot{m}_{t-1} + m_{t-1}, \\
\dot{w}_{t} &= \lambda_{t-1} \dot{w}_{t-1} + w_{t-1}, \\
\label{eqt-dLt}
\Delta(L_{t}, {\boldsymbol{\lambda}_{t-2}}) &= 2 (\hat{Y}_{t-1} - Y_{t}) \left( \frac{\dot{m}_{t-1} - \dot{w}_{t-1} \hat{Y}_{t-1}  }{w_{t-1}} \right).
\end{align}
In \citet{Bodenham2016}, the authors suggest to scale $\Delta(L_{t}, {\boldsymbol{\lambda}_{t-2}})$ in (\ref{eqt-lambda}) to $\Delta(L_{t}, {\boldsymbol{\lambda}_{t-2}})/\hat{\sigma}^{2}$ ($\hat{\sigma}^{2}$ is the sample variance that can be estimated during a burn-in period). The reason is that large variation in $Y_{i}$'s will force the forgetting factors, $\lambda_{1}, \cdots, \lambda_{t}$, computed via (\ref{eqt-lambda}) to be either 0 or 1. However, in this paper, we are only interested in Bernoulli rewards, which means that the variation in $Y_{i}$'s is less than 1, so it is not essential to devise an elaborate scaling scheme.

In addition to the mean, we may make use of an adaptive estimate of the variance. The adaptive forgetting factor variance is defined as:
\begin{align}
\label{eqt-AFFvariance}
s_{t}^{2} = \frac{1}{v_{t}} \sum_{i=1}^{t} \left(\prod_{p=i}^{t-1} \lambda_{p} \right) (Y_{i} - \hat{Y}_{t})^{2},
\end{align}
where $v_{t} = w_{t} \left(1- \frac{k_{t}}{(w_{t})^2} \right)$, $k_t = \sum_{i=1}^{t} \left(\prod_{p=i}^{t-1} \lambda_{p}^{2} \right)$, and $v_{t}$ is selected to make $\mathbb{E}[s_{t}^{2}] = \text{Var}[Y_{i}]$ when $Y_{i}$'s are i.i.d.
Note here we choose the same adaptive forgetting factor for mean and variance for convenience, though other formulations are possible.  One can use a separate  adaptive forgetting factor for the variance if needed. Again, $s_{t}^{2}$ can be computed recursively via the following equations:
\begin{align}
s_{t}^{2} & =  \frac{1}{v_{t}} \left[ \left(\lambda_{t-1} v_{t-1} \right) s^{2}_{t-1} + \left(\frac{w_{t} - 1}{w_{t} }\right) (\hat{Y}_{t-1} - Y_{t})^2 \right], \\
k_{t} & = \lambda_{t-1}^{2} k_{t-1} + 1, \\
v_{t} & = w_{t} \left(1- \frac{k_{t}}{(w_{t})^2} \right).
\end{align}


\subsection{Updating Estimation when not Selecting}
\label{sec-missing}

In the MAB setting, we have at least two arms, and for each arm, we will construct an AFF estimator. However, we can only observe one arm at a time. This means that the estimations and intermediate quantities of an unobserved arm will retain their previous values, that is, if arm $a$ is not observed at time $t$, 
\begin{align*}
\hat{Y}_{t}(a) & = \hat{Y}_{t-1}(a), \\
s_{t}^{2}(a) & = s_{t-1}^{2}(a), \\
\lambda_{t}(a) & = \lambda_{t-1}(a), \\
m_{t}(a) &= m_{t-1}(a), \\
w_{t}(a) &= w_{t-1}(a), \\
k_{t}(a) &= k_{t-1}(a).
\end{align*}

Not being able to update estimators sets more challenges in dynamic cases. In static cases, the sample mean estimator will converge quickly to the expected reward with a few observations, and therefore it has little effect if the arm is not observed further. However, in dynamic cases, even if the estimator tracks the expected reward perfectly at a given moment, its precision may deteriorate quickly once it stops getting new observations. Therefore, it is more challenging to balance exploration and exploitation in dynamic cases. 
In the following section, we introduce how to modify the popular MAB methods using estimators $\hat{Y}_{t}$ and $s_{t}^{2}$ as well as the intermediate quantities $m_{t}$, $w_{t}$ and $k_{t}$. In some cases (UCB and TS), these modifications will result in an increase of the uncertainty of estimations of arms that have not been observed for a while. This appears as a discounting of $m_{t}$, $w_{t}$, and $k_{t}$ as shown later in Section~\ref{sec-discount}.

 The tuning parameter in AFF estimation is the step size $\eta$ used in (\ref{eqt-lambda}); its choice may affect the performance of estimation, and thus affect the performance of our AFF-deployed MAB algorithms that will be introduced in Section~\ref{sec-selection}. We examine empirically the influence of $\eta$ on these algorithms in Section \ref{sec-initialisation}. 

\section{Action Selection}
\label{sec-selection}

Having discussed how to track the expected reward of arms in the previous section, we now move on to methods for the selection step. 
We will consider three of the most popular methods: $\epsilon$-Greedy \citep{Watkins1989}, UCB \citep{Auer2002c} and TS \citep{Thompson1933}. They are easy to implement and computationally efficient. Moreover, they have good performance in numerical evaluations \citep{Chapelle2011, Kuleshov2014, Vermorel2005}. Each of these methods uses a different mechanism to balance the exploration-exploitation trade-off. Deploying AFF in these methods, we propose a new family of MAB algorithms for dynamic Bernoulli bandits that are denoted with the prefix AFF- to emphasise the use of AFF in estimation. We call our new algorithms as AFF-$d$-Greedy, AFF-UCB1/AFF-UCB2, and AFF-TS/AFF-OTS corresponding to different types of modifications.

In the literature of dynamic bandits, many approaches attempted to improve the performance in standard methods by choosing an estimator that uses the reward history wisely.
\citet{Koulouriotis2008} applied exponentially-weighted average estimation in $\epsilon$-Greedy. \citet{Kocsis2006} introduced the discounted UCB method (it was also called D-UCB in \citealp{Garivier2011}) which used a fixed discounting factor in estimation. \citet{Garivier2011} proposed the Sliding Window UCB (SW-UCB) algorithm where the reward history used for estimation is restricted by a window.  The Dynamic Thompson Sampling (DTS) algorithm applied a bound on the reward history used for updating the hyperparameters in posterior distribution of $\mu_{t}$ \citep{Gupta2011}. \citet{Raj2017} applied a fixed forgetting factor in Thompson sampling and proposed discounted Thompson sampling (dTS). These sophisticated algorithms require accurate tuning of some input parameters, which relies on knowledge of the model/behaviour of $\mu_{t}$. For example, computing the window size of SW-UCB, or the discounting factor of D-UCB \citep{Garivier2011} requires knowing the number of switch points (i.e., number of times that the optimal arm switches). While the idea behind our AFF MAB algorithms is similar, our approaches automate the tuning of the key parameters (i.e., the forgetting factors), and require only little effort to tune the higher level parameter $\eta$ in (\ref{eqt-lambda}). Moreover, we use the AFF technique to guide the tuning of the key parameter in the DTS algorithm, which will be discussed later in this section. 


In what follows, we discuss each AFF-deployed method separately. We briefly review the basics of each method and refer the reader to the references for more details. In addition, we will continue to use notations like $Y_{t}$ instead of $Y_{t}(a)$ when clear. In all the AFF MAB algorithms we propose below, we will use a very short initialisation (or burn-in) period for the initial estimations. Normally, the length of the burn-in period is $|\mathcal{A}|$, that is, selecting each arm once; for the algorithms that require estimate of variance, we use a longer burn-in period by selecting each arm $M$ times.

\subsection{$\epsilon$-Greedy}

$\epsilon$-Greedy \citep{Watkins1989} is the simplest method for the static MAB problem. The expected reward of an arm is estimated by its sample mean, and a fixed parameter $\epsilon \in (0, 1)$ is used for selection. At each time step, with probability $\epsilon$, the algorithm selects an arm uniformly to explore, and with probability $1-\epsilon$, the arm with the highest estimated reward is picked. $\epsilon$-Greedy is simple and easy to implement, which makes it appealing for dynamic bandits. However, it can have two main issues:
first, the sample average is not ideal for tracking the moving reward;
second, the parameter $\epsilon$ is the key to balancing the exploration-exploitation dilemma, but it is challenging to tune as an optimal strategy in dynamic environments may require varying $\epsilon$ over time. Indeed, all numerical studies in Section~\ref{sec-exps} support this statement.



\begin{algorithm}[H]
\caption{AFF-$d$-Greedy}
\label{algo-AFF-d-Greedy}
\begin{algorithmic}
\REQUIRE $d \in (0, 1)$; $\eta \in (0,1)$.
\STATE \textbf{Initialisation:} play each arm once.
\FOR{$t = |\mathcal{A}|+1, \cdots, T$}
\STATE find $a_{t} = \argmax_{a' \in \mathcal{A}} \hat{Y}_{t-1}(a')$;
\IF{$|\lambda_{t-1}(a_{t}) - \lambda_{t-2}(a_{t})| \geq d$}
\STATE choose $a_{t}$ uniformly from $\mathcal{A}$;
\ENDIF
\STATE select arm $a_t$ and observe reward $Y_{t}(a_{t})$;
\STATE update $\hat{Y}_{t}(a_{t})$.
\ENDFOR
\end{algorithmic}
\end{algorithm}    

In Algorithm~\ref{algo-AFF-d-Greedy}, we propose the AFF-$d$-Greedy algorithm to overcome the above weaknesses. In the algorithm, we use the AFF mean $\hat{Y}_{t}$ from (\ref{eqt-AFFmean}) to estimate the expected reward. This estimator can respond quickly to changes, that is, 
for an arm that is frequently observed, it can closely follow the underlying reward; 
for an arm that is not observed for a long time, the estimator can capture $\mu_{t}$ quickly once the arm is selected again. At each time step, we first identify the arm with the highest AFF mean; if the absolute difference between this arm's last two forgetting factors is smaller than $d$, we select it; otherwise, we select an arm from $\mathcal{A}$ uniformly. A threshold $d \in (0,1)$ is used to balance exploration and exploitation. Tuning $d$ is easier than $\epsilon$ as it is related to the step size $\eta$ used in (\ref{eqt-lambda}). This was confirmed in a large number of simulations. For Bernoulli dynamic bandits, we suggest to set $d \approx \eta$. 

We use the forgetting factors $\lambda_{t}$ $(t=1, 2, \cdots)$ in the decision rule 
as their magnitudes indicate the variability of the reward stream. For example, if $\lambda_{t}$ is close to zero, it can be interpreted as a sudden change occurring at time $t$, and if close to 1, it indicates that the reward stream is stable at time $t$. To understand the decision rule better, we illustrate it using two examples.
\begin{enumerate}
\item Variable arm example: let us say arm $\hat{a}$ was selected at time $t-1$, and at time $t$, arm $\hat{a}$ has the highest estimated reward and $|\lambda_{t-1}(\hat{a}) - \lambda_{t-2}(\hat{a})| < d$. By the decision rule, the algorithm will select this arm again.
We are interested in two cases: first, both $\lambda_{t-1}(\hat{a})$ and $\lambda_{t-2}(\hat{a})$ are close to 1; second, both $\lambda_{t-1}(\hat{a})$ and $\lambda_{t-2}(\hat{a})$ are close to 0. It is easy to understand why the algorithm select it in the first case, as the arm is currently stable and it has the highest estimated reward. In the second case, $\mu_{t}(\hat{a})$ seems variable in the past two steps. Even if $\mu_{t}(\hat{a})$ had kept moving down (that is, the worst possibility), the estimated reward would have fallen as well, since arm $\hat{a}$ still has the highest estimated reward, Algorithm~\ref{algo-AFF-d-Greedy} will select it.

\item Idle arm example: let us say arm $\hat{a}$ has the highest estimated reward at time $t$, and it was not selected at $t-1$. By the decision rule, Algorithm~\ref{algo-AFF-d-Greedy} will select this arm since $|\lambda_{t-1}(\hat{a}) - \lambda_{t-2}(\hat{a})| = 0$. 
\end{enumerate}
From these examples, we can see that exploration and exploitation are balanced in a way that takes into account the variability in the estimation procedure rather than by simply flipping a coin. 
Loosely speaking, when variability appears in a dominant arm, one could expect that sufficient time has passed in order to trigger exploration. 
We do not claim this is a powerful algorithm. Indeed, we are trying to develop an analogue to $\epsilon$-Greedy that responds to dynamics.

\subsection{Upper Confidence Bound}
\label{sec-UCB}
Another type of algorithm uses \textit{upper confidence bounds} for selection. 
The idea is that, instead of the plain sample average, an exploration bonus is added to account for the uncertainty in the estimation, and the arm with highest potential of being optimal will be selected. This exploration bonus is typically derived using concentration inequalities \citep[e.g.,][]{Hoeffding1963}. The UCB1 algorithm introduced by \citet{Auer2002c} is a classic method. In latter works, UCB1 was often called simply UCB. For any reward distribution that is bound in [0,1], the UCB algorithm picks the arm which maximise the quantity $\bar{Y}_{t} + \sqrt{\frac{2 \log t}{N_{t}}}$, where $\bar{Y}_{t}$ is the sample average and $N_{t}$ is the number of times this arm was played up to time $t$. The exploration bonus $\sqrt{\frac{2 \log t}{N_{t}}}$ was derived using the Chernoff-Hoeffding bound. It is proved that the UCB algorithm achieves logarithmic regret uniformly over time \citep{Auer2002c}. 

For better adaptation in dynamic environments, we replace $\bar{Y}_{t}$ with $\hat{Y}_{t}$, and modify the upper bound accordingly as:
\begin{align}
\hat{Y}_{t} + \sqrt{\frac{- \log (0.05)}{2 (w_{t})^2/ k_{t} }},
\end{align}
where $w_{t}$ and $k_{t}$ are intermediate quantities related to AFF estimation (see Section \ref{sec-AdaEst}).
The modified exploration bonus $\sqrt{\frac{- \log (0.05)}{2 (w_{t})^2/ k_{t} }}$ is derived via Hoeffding's inequality in a similar way to the derivation of UCB (see Appendix \ref{app-AFFUCB} for details). However, for Arm $\hat{a}$ that is not selected for a while, its upper bound will be static since $\hat{Y}_{t}(\hat{a})$, $w_{t}(\hat{a})$ and $k_{t}(\hat{a})$ do not change. As a consequence, Arm $\hat{a}$ will only be selected if the arm with current highest upper bound drops below it. 
This is not desirable since one needs to be gradually more agnostic towards the belief on its expected reward $\mu_{t}(\hat{a})$ in a dynamic environment.
Inflating the uncertainty can be implemented in different ways, either by adjusting the upper bound or discounting the quantities used in AFF estimation. This results in two possible algorithms: AFF-UCB1 and AFF-UCB2.

\subsubsection{Adjusting the Upper Bound for Unselected Arms}

The upper bound for selection at time $t+1$ takes the form $\hat{Y}_{t} + B_{t}$, and $B_{t}$ is:
\begin{align}
\label{seq-AFFUCB-bound}
B_{t} = \sqrt{\frac{- \log (0.05)}{2 (w_{t})^2/ k_{t} }} \mathbb{I}(t - t_{last}=0) + \sqrt{\frac{s^{2}_{t}}{w_{t} }} (t - t_{last})^{1/|\mathcal{A}|},
\end{align}
where $t_{last}$ is the last time instant that the arm was observed, and $s_{t}^{2}$ is the AFF variance defined in (\ref{eqt-AFFvariance}). We propose AFF-UCB1 in Algorithm~\ref{algo-AFF-UCB}.

The exploration bonus $B_{t}$ is a combination of two components. It can be interpreted by considering two cases:
\begin{enumerate}
\item if an arm was observed at the previous time step, $t$ (i.e., $t - t_{last}=0$), its exploration bonus is $B_{t} = \sqrt{\frac{- \log (0.05)}{2 (w_{t})^2/ k_{t} }}$, which is analogous to that of UCB;
\item if an arm was not observed at the previous time step, its exploration bonus is constructed as $B_{t} = \sqrt{\frac{s^{2}_{t}}{w_{t} }} (t - t_{last})^{1/|\mathcal{A}|}$. This exploration bonus inflates with the time of being idle in order to compensate the uncertainty caused by not being observed. Note here $B_{t}$ decreases with the number of arms, $|\mathcal{A}|$. This makes use of the fact that as $|\mathcal{A}|$ increases, the population of arms will ``fill'' more the reward space and more opportunities will arise for picking high reward arms.
\end{enumerate}

\begin{algorithm}[H]
\caption{AFF-UCB1}
\label{algo-AFF-UCB}
\begin{algorithmic}
\REQUIRE $\eta \in (0,1)$.
\STATE \textbf{Initialisation:} play each arm $M$ times.
\FOR{$t = M |\mathcal{A}|+1, \cdots, T$}
\STATE for all $a \in \mathcal{A}$, compute $B_{t-1}(a)$ according to (\ref{seq-AFFUCB-bound});
\STATE find $a_{t} = \arg\max_{a' \in \mathcal{A}}  \left( \hat{Y}_{t-1}(a') + B_{t-1}(a')  \right) $;
\STATE  select arm $a_t$ and observe reward $Y_{t}(a_{t})$;
\STATE update $\hat{Y}_{t}(a_{t})$, $w_{t}(a_{t})$, $k_{t}(a_{t})$, $s_{t}^{2}(a_{t})$, and $t_{last}(a_{t})$.
\ENDFOR 
\STATE Note here, we use a longer burn-in period since we need to initialise the estimation of data variance. In the simulation study in Section~\ref{sec-exps}, we choose $M=10$.
\end{algorithmic}
\end{algorithm}

\subsubsection{Discounting of $m_{t}$, $w_{t}$, and $k_{t}$ for Unselected Arms}
\label{sec-discount}
Instead of deliberately adding inflation as in (\ref{seq-AFFUCB-bound}), alternatively we can discount directly on the intermediate quantities  $m_{t}$, $w_{t}$, and $k_{t}$ and use them in the decision. Letting the updating of $m_{t}$, $w_{t}$, and $k_{t}$ remain the same as in Section~\ref{sec-AdaEst}, we introduce quantities $\tilde{m}_{t}$, $\tilde{w}_{t}$, and $\tilde{k}_{t}$ which are computed as:
\begin{align}
\label{eqt-tilde-mt}
\tilde{m}_{t} &= (\lambda_{t})^{\frac{t-t_{last}}{|\mathcal{A}|}} m_t, \\
\label{eqt-tilde-wt}
\tilde{w}_{t} &= (\lambda_{t})^{\frac{t-t_{last}}{|\mathcal{A}|}} w_t, \\
\label{eqt-tilde-kt}
\tilde{k}_{t} &= (\lambda_{t}^{2})^{\frac{t-t_{last}}{|\mathcal{A}|}} k_t,
\end{align}
where $t_{last}$ is the last time instant that the arm was selected. 

If an arm is selected, the two sets of quantities are identical, i.e., $\tilde{m}_{t}=m_{t}$, $\tilde{w}_{t}=w_{t}$, and $\tilde{k}_{t}=k_{t}$. 
If an arm is not selected, $m_{t}$, $w_{t}$, and $k_{t}$ are discounted according to the forgetting factor obtained when it was selected last time (note that the forgetting factor $\lambda_{t}$ of an unselected arm remains the same as $\lambda_{t_{last}}$).

Using $\tilde{w}_{t}$, and $\tilde{k}_{t}$, we modify the upper bound for selection at time $t+1$ to $\hat{Y}_{t} + \tilde{B}_{t}$, where 
\begin{align}
\label{seq-AFFUCB-bound2}
\tilde{B}_{t} = \sqrt{\frac{- \log (0.05)}{2 (\tilde{w}_{t})^2/ \tilde{k}_{t} }},
\end{align}
and this results in AFF-UCB2 in Algorithm~\ref{algo-AFF-UCB2}. 

\begin{algorithm}[H]
\caption{AFF-UCB2}
\label{algo-AFF-UCB2}
\begin{algorithmic}
\REQUIRE $\eta \in (0,1)$.
\STATE \textbf{Initialisation:} play each arm once.
\FOR{$t = |\mathcal{A}|+1, \cdots, T$}
\STATE for all $a \in \mathcal{A}$, compute $\tilde{B}_{t-1}(a)$ according to (\ref{seq-AFFUCB-bound2});
\STATE find $a_{t} = \arg\max_{a' \in \mathcal{A}}  \left( \hat{Y}_{t-1}(a') + \tilde{B}_{t-1}(a')  \right) $;
\STATE  select arm $a_t$ and observe reward $Y_{t}(a_{t})$;
\STATE update $\hat{Y}_{t}(a_{t})$, $w_{t}(a_{t})$, $k_{t}(a_{t})$, and $t_{last}(a_{t})$;
\STATE  update $\tilde{w}_{t}(a)$ and $\tilde{k}_{t}(a)$ for all $a \in \mathcal{A}$.
\ENDFOR 
\end{algorithmic}
\end{algorithm}

\subsection{Thompson Sampling}

Recently, researchers \citep[e.g.,][]{Scott2015} have given more attention to the Thompson Sampling (TS) method which can be dated back to \citet{Thompson1933}. It is an approach based on Bayesian principles. A usually conjugate prior is assigned to the expected reward of each arm at the beginning, and the posterior distribution of the expected reward is sequentially updated through successive  arm selection. A decision rule is constructed using this posterior distribution. At each round, a random sample is drawn from the posterior distribution of each arm, and the arm with the highest sample value is selected. 

For the static Bernoulli bandit, following the approach of \citet{Chapelle2011}, it is convenient to choose the Beta distribution, $Beta(\alpha_{0}, \beta_{0})$, as a prior. The posterior distribution is then $Beta(\alpha_{t}, \beta_{t})$ at time $t$, and the parameters $\alpha_{t}$ and $\beta_{t}$ can be updated recursively as follows:
if an arm is selected at time $t$, 
\begin{align}
\label{eqt-at-TS}
\alpha_{t} &= \alpha_{t-1} + Y_{t}, \\
\label{eqt-bt-TS}
\beta_{t} &= \beta_{t-1} +  1 - Y_{t};
\end{align}
otherwise,
\begin{align}
\label{eqt-at2-TS}
\alpha_{t} &= \alpha_{t-1}, \\
\label{eqt-bt2-TS}
\beta_{t} &= \beta_{t-1}.
\end{align}

The simplicity and effectiveness in real applications \citep{Scott2015} make TS a good candidate for dynamic bandits. However, it has similar issues in tracking $\mu_{t}$ as in $\epsilon$-Greedy and UCB. For illustration, assume an arm is observed all the time, then one can re-write the recursions in (\ref{eqt-at-TS})-(\ref{eqt-bt-TS}) as:
\begin{align*}
\alpha_{t} & = \alpha_{0} + \sum_{i=1}^{t}Y_{i}, \\
\beta_{t} & = \beta_{0} + \sum_{i=1}^{t}1 - \sum_{i=1}^{t}Y_{i}.
\end{align*}
As a result, the posterior distribution $Beta(\alpha_{t}, \beta_{t})$ keeps full memory of all the past observations, making posterior inference less responsive to observations near time $t$.

To modify the above updating, we use the quantities $\tilde{m}_{t}$ and $\tilde{w}_{t}$ from (\ref{eqt-tilde-mt})-(\ref{eqt-tilde-wt}), and have
\begin{align}
\label{eqt-at-AFFTS}
\alpha_{t} &= \alpha_{0} + \tilde{m}_{t} , \\
\label{eqt-bt-AFFTS}
\beta_{t} &=  \beta_{0} + \tilde{w}_{t} - \tilde{m}_{t}.
\end{align}

Similar to the AFF-UCB2 algorithm, the exploration of unselected arms is boosted by using the discounted quantities $\tilde{m}_{t}$ and $\tilde{w}_{t}$. To be more specific, the posterior distribution is flattened for an unselected arm, and the longer the arm is unselected, the further its posterior distribution is flattened. 
With updates (\ref{eqt-at-AFFTS})-(\ref{eqt-bt-AFFTS}), we propose in Algorithm~\ref{algo-AFF-TS} the AFF-TS algorithm for dynamic Bernoulli bandits. We should mention here that \citet{Raj2017} proposed a similar algorithm called discounted Thompson sampling where the authors discount the hyper-parameters using a fixed forgetting factor.


\begin{algorithm}[H]
\caption{AFF-TS for Dynamic Bernoulli Bandits}
\label{algo-AFF-TS}
\begin{algorithmic}
\REQUIRE $Beta(\alpha_{0}(a), \beta_{0}(a))$ for all $a \in \mathcal{A}$; $\eta \in (0,1)$.
\STATE \textbf{Initialisation:} play each arm once.
\FOR{$t = |\mathcal{A}|+1, \cdots, T$}
\STATE for all $a \in \mathcal{A}$, draw a sample $x(a)$ from $Beta(\alpha_{t-1}(a), \beta_{t-1}(a))$;
\STATE find $a_{t} = \arg\max_{a' \in \mathcal{A}} x(a')$;
\STATE  select arm $a_t$ and observe reward $Y_{t}(a_{t})$;
\STATE  update $\alpha_{t}(a)$ and $\beta_{t}(a)$ according to (\ref{eqt-at-AFFTS})-(\ref{eqt-bt-AFFTS}) for all $a \in \mathcal{A}$.
\ENDFOR
\end{algorithmic}
\end{algorithm}

\subsubsection{Optimistic Thompson Sampling}
\label{sec-OTS}
We now look at some popular extensions of TS.
\citet{May2012a} introduced the \textit{optimistic} version of Thompson sampling called Optimistic Thompson Sampling (OTS), where the drawn sample value is replaced by its posterior mean if the former is smaller. That is to say, for each arm, the score used for decision will never be smaller than the posterior mean. OTS boosts further the exploration of highly uncertain arms compared to TS, as OTS increases the probability of getting a high score for arms with high posterior variance. 

However, OTS has the same problem as TS when applied to a dynamic problem, that it uses the full reward history to update the posterior distribution. We propose the AFF version of OTS in Algorithm~\ref{algo-AFF-OTS}.

\begin{algorithm}[H]
\caption{AFF-OTS for Dynamic Bernoulli Bandits}
\label{algo-AFF-OTS}
\begin{algorithmic}
\REQUIRE $Beta(\alpha_{0}(a), \beta_{0}(a))$ for all $a \in \mathcal{A}$; $\eta \in (0,1)$.
\STATE \textbf{Initialisation:} play each arm once.
\FOR{$t = |\mathcal{A}|+1, \cdots, T$}
\STATE for all $a \in \mathcal{A}$, draw a sample $x(a)$ from $Beta(\alpha_{t-1}(a), \beta_{t-1}(a))$, \\ and replace $x(a)$ with $\frac{\alpha_{t-1}(a)}{\alpha_{t-1}(a) + \beta_{t-1}(a)}$ if $x(a)$ is smaller;
\STATE find $a_{t} = \arg\max_{a' \in \mathcal{A}} x(a')$;
\STATE  select arm $a_t$ and observe reward $Y_{t}(a_{t})$;
\STATE  update $\alpha_{t}(a)$ and $\beta_{t}(a)$ according to (\ref{eqt-at-AFFTS})-(\ref{eqt-bt-AFFTS}) for all $a \in \mathcal{A}$.
\ENDFOR
\end{algorithmic}
\end{algorithm}

\subsubsection{Tuning Parameter $C$ in Dynamic Thompson Sampling}
\label{sec-TuneC}

The Dynamic Thompson Sampling (DTS) algorithm was introduced by \citet{Gupta2011} specifically for solving the dynamic Bernoulli bandit problem of interest here. The DTS algorithm uses a pre-determined threshold $C$ in updating the posterior parameters $\alpha_{t}$ and $\beta_{t}$ while using the standard Thompson sampling technique for arm selection. For the arm that is selected at time $t$, if $\alpha_{t-1} + \beta_{t-1} < C$, the posterior parameters are updated via (\ref{eqt-at-TS})-(\ref{eqt-bt-TS}); otherwise when $\alpha_{t-1} + \beta_{t-1} \geq C$, 
\begin{align*}
\alpha_{t} &= \left(\alpha_{t-1} + Y_{t} \right)\frac{C}{C+1}, \\
\beta_{t} &= \left( \beta_{t-1} + 1 - Y_{t} \right) \frac{C}{C+1}.
\end{align*}

To understand, let $\hat{\mu}_{t}$ denote the posterior mean, and assume an arm is observed all the time. Say at time $s$ the arm achieves the threshold, i.e., $\alpha_{t} + \beta_{t} = C$ for $t = s$ and onward. 
Following (17)-(21) in \citet{Gupta2011}, 
\begin{align}
\label{eqt-DTSmean}
\hat{\mu}_{t}  &= \left(1 - \frac{1}{C+1} \right) \hat{\mu}_{t-1} + \frac{1}{C+1} Y_{t},
\end{align}
which is a weighed average of $\hat{\mu}_{t-1}$ and the observation $Y_{t}$. The recursion of $\hat{\mu}_{t}$ is similar to the EWMA scheme \citep{Roberts1959}. Essentially, the DTS algorithm uses the threshold $C$ to bound the total amount of reward history used for updating the posterior distribution. Once the threshold is achieved, the algorithm puts more weight on newer observations.


Although it was demonstrated in \citet{Gupta2011} that the DTS algorithm has the ability to track the changes in the expected reward, the performance of the algorithm is very sensitive to the choice of $C$. In our numerical  simulations (see Section~\ref{sec-exp-AFFDTS}), we found that the performance of the DTS algorithm varies widely with different $C$ values. However, in \citet{Gupta2011}, the authors did not provide tuning methods for $C$. To address this issue, we propose below two different ways to tune $C$ adaptively at each time step using AFF estimation (AFF-DTS1~\&~2 resp.). 

\paragraph{AFF-DTS1}
From the numerical results in \citet[sect. IV.C]{Gupta2011}, the optimal $C$ is related to the the speed of change of $\mu_{t}$. This motivates us to tune $C$ according to the variance of rewards obtained. We can use the AFF variance, $s_{t}^{2} $, defined in (\ref{eqt-AFFvariance}) as an estimate of the reward variance.
One option is to use $C_{t} \propto 1/s_{t}^{2} $; since high $s_{t}^{2} $ indicates more dynamics in $\mu_{t}$, a shorter reward history is required.
For example in the numerical examples in Section~\ref{sec-exp-AFFDTS}, we will use $C_{t} = \frac{4}{s^{2}_{t}} - 1$.

\paragraph{AFF-DTS2}
Another way to set $C_t$ is based on the similarity of the posterior mean in DTS and the AFF mean introduced in Section~\ref{sec-AdaEst}. In particular, in (\ref{eqt-DTSmean}) the posterior mean is given by:
$$\hat{\mu}_{t} = (1 - \frac{1}{C+1}) \hat{\mu}_{t-1} + \frac{1}{C+1} Y_{t}. $$
Using (\ref{eqt-AFFmean})-(\ref{eqt-wt}), one can re-write the the AFF mean as:
$$\hat{Y}_{t} = \left(1 - \frac{1}{w_{t}} \right) \hat{Y}_{t-1} + \left(\frac{1}{w_{t}} \right) Y_{t}.$$
Therefore, at each time step $t$, we can set $C_{t} =w_{t} - 1$.

\section{Numerical Results}
\label{sec-exps}
In this section, we illustrate the performance improvements on $\epsilon$-Greedy, UCB, and TS using AFFs. 
The first simulation study examines how the algorithms behave for a small number of changes.
We then consider two different dynamic scenarios for the expected reward $\mu_{t}$: abruptly changing and drifting. For the abruptly changing scenario, instead of manually setting up change points in $\mu_{t}$ as in \citet{Yu2009} and \citet{Garivier2011}, we set up change-point instants for an arm by an exponential clock (see Section \ref{sec-AbruptChange}). In the drifting scenario, the evolution of the expected reward $\mu_{t}$ is driven by a random walk in the interval (0,1), and we use two different models: the first model is inspired by \citet{Slivkins2008} where $\mu_{t}$ is modelled by a random walk with reflecting bounds; the second model is to use a transformation function on a random walk. For each scenario, we test the performance with 2, 50 and 100 arms; the two-armed examples are used for the purpose of illustration, and the latter examples (50 and 100 arms) are used to evaluate the performance with a large number of arms. Results for D-UCB and SW-UCB \citep{Garivier2011} are also reported in the studies mentioned above.
We further demonstrate the robustness of the AFF MAB algorithms to tuning, specifically, sensitivity to the step size, $\eta$. Finally, we use a two-armed example to show that the modified DTS algorithms, i.e., AFF-DTS1 and AFF-DTS2, can reduce the performance sensitivity of DTS to the input parameter $C$.

The tuning parameters in the algorithms are initialised as follows. For the $\epsilon$-Greedy method, we evaluate over a grid of choice of $\epsilon$, $\epsilon \in (0.1, \cdots, 0.9)$, and report performance for the \textit{best} choice. While this is unrealistic, it provides a useful benchmark. We use step size $\eta = 0.001$ for all AFF MAB algorithms. For AFF-$d$-Greedy, the threshold $d$ is set as $d=\eta$. For all Thompson sampling based algorithms, we use $Beta(2, 2)$ as the prior. According to \citet{Garivier2011}, the fixed discounting factor $\lambda$ in D-UCB is set to $\lambda = 1-(4)^{-1}\sqrt{ \Upsilon_{T}/T}$, and the window size $W$ in SW-UCB is set to $W = 2\sqrt{T\log(T )/ \Upsilon_{T}}$, where $\Upsilon_{T}$ is the number of switch points during the total $T$ rounds.

\subsection{Performance for a Small Number of Changes}

\begin{figure}[H]
    \centering
    \subfloat[$\mu_t$ against $t$.]
    {\includegraphics[width=0.55\textwidth, height=0.2\textheight]{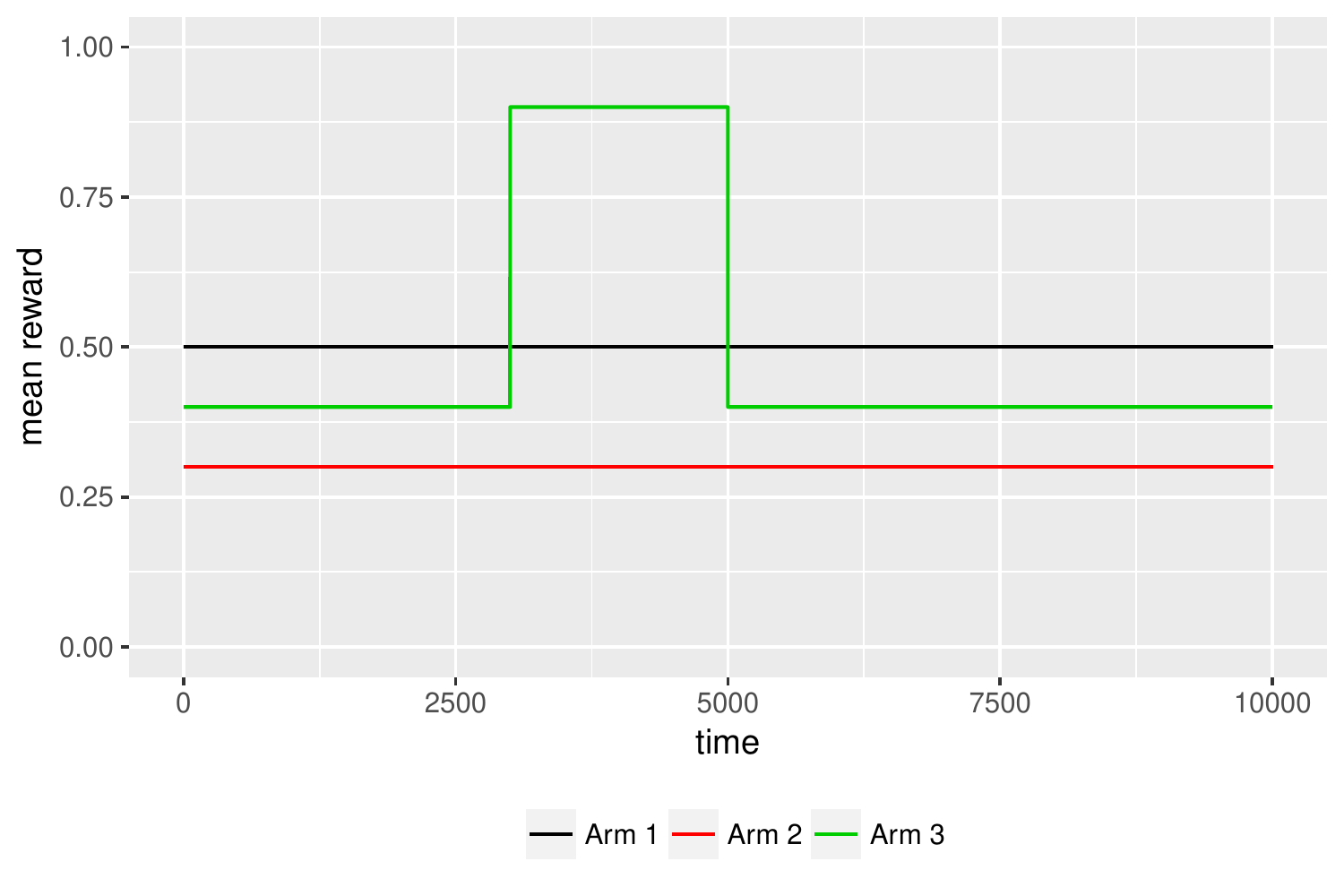}}\hfill
    \subfloat[Cumulative regret.]
    {\includegraphics[width=0.45\textwidth, height=0.2\textheight]{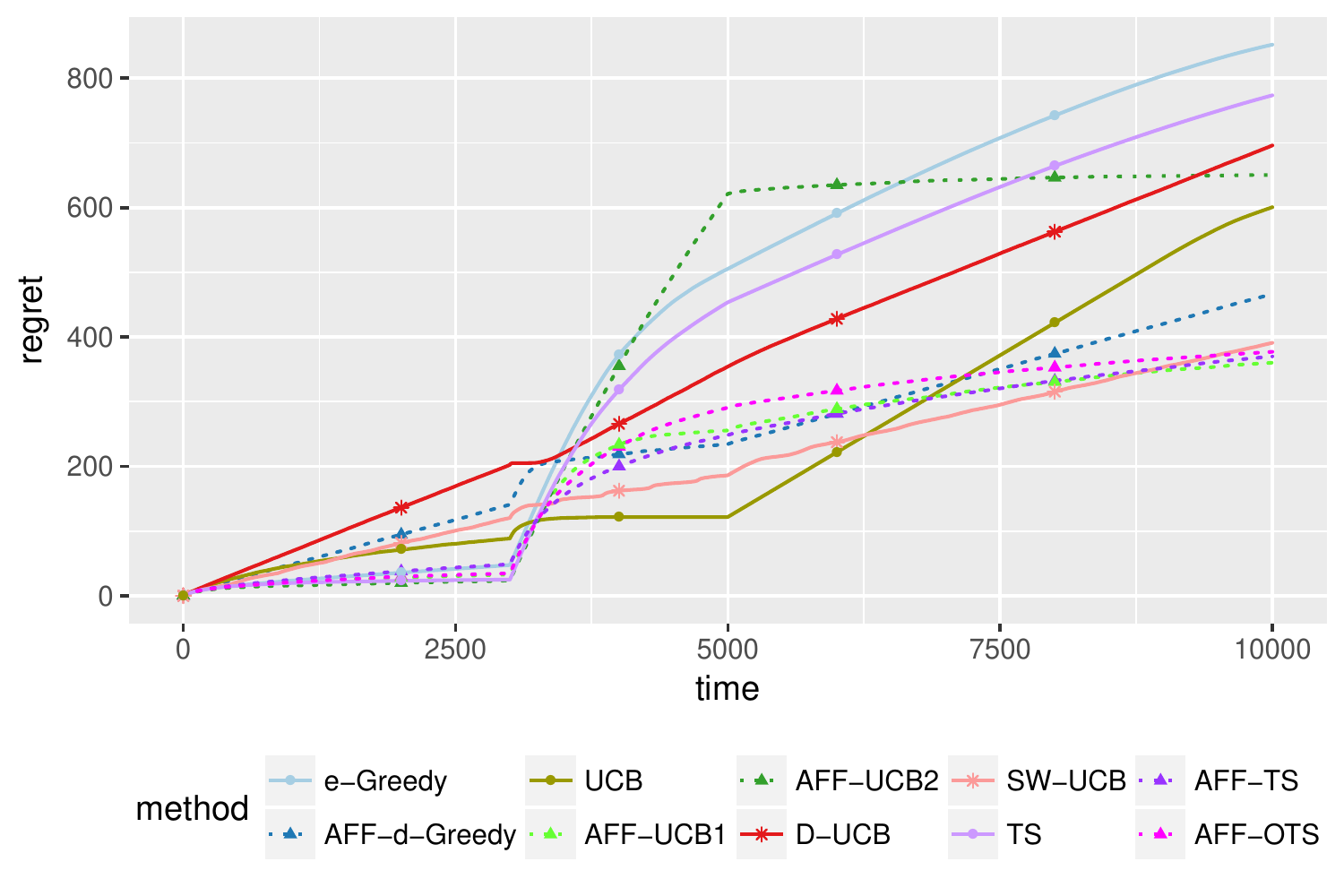}}\hfill
    \subfloat[Boxplot of total regret at $T=10,000$.]
    {\includegraphics[width=0.45\textwidth, height=0.2\textheight]{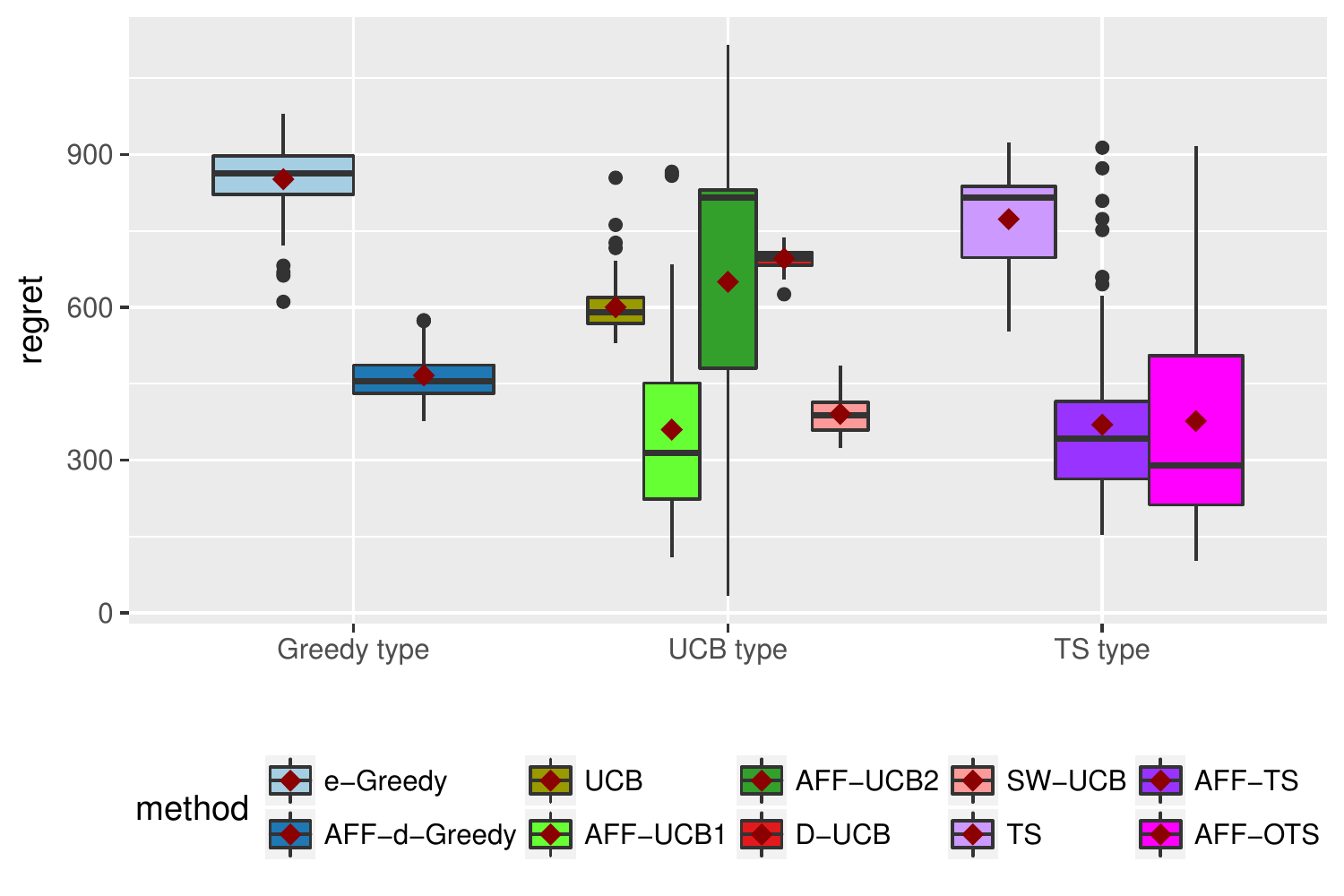}}\hfill
    \subfloat[Percentage of correct selections at every time step over 100 replications.]
    {\includegraphics[width=0.9\textwidth, height=0.22\textheight]{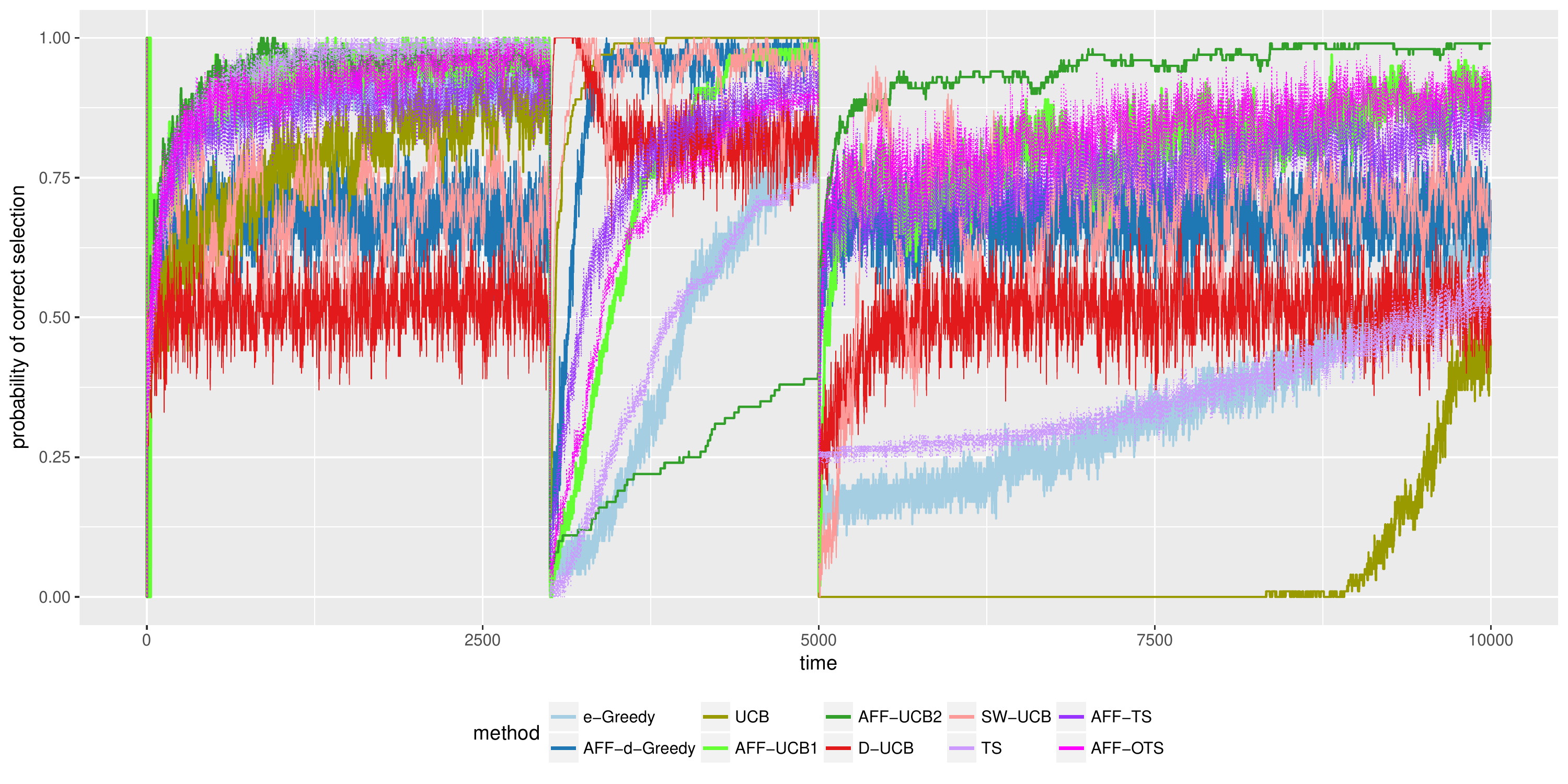}}\hfill
    
    \caption{Performance of different algorithms in the case of small number of changes.}
    \label{plot-SmallChange} 
\end{figure}

We illustrate how our algorithms behave when an (initially) inferior arm becomes optimal using the example displayed in Figure~\ref{plot-SmallChange}(a). The total length of the simulated experiment is $T=10,000$. The expected reward $\mu_{t}$ of Arm~1 and Arm~2 are 0.5 and 0.3 respectively for $t=1, \cdots, 10000$. Two changes of Arm~3 occur at $t = 3,000$ ($\mu_{t}$ jumped from 0.4 to 0.8) and $t=5,000$ ($\mu_{t}$ dropped back to 0.4).  This setting is the same to the first simulation example in \citet{Garivier2011}.

Figure~\ref{plot-SmallChange}(b) displays the cumulative regret of all algorithms (the results are averaged over 100 independent replications), and Figure~\ref{plot-SmallChange}(c) shows boxplots of the total regret at $t=10,000$. It can be seen that the performance of AFF-$d$-Greedy, AFF-TS, and AFF-OTS are much better than standard methods. Among UCB type methods, AFF-UCB1 improves the standard method UCB, and its performance is similar to SW-UCB; AFF-UCB2 and D-UCB perform slightly worse than UCB. In Figure~\ref{plot-SmallChange}(d), we present the percentage of correct arm selections over the 100 replications for each time step. As can be seen, our AFF MAB algorithms (except AFF-UCB2) have the ability to \textit{detect and respond} fairly quickly to the changes at $t = 3000$ and $5000$.

\subsection{Performance for Different Dynamic Models}
We first use two-armed examples to compare the performance of AFF-$d$-Greedy, AFF-UCB1/AFF-UCB2, and AFF-TS/AFF-OTS to the standard methods $\epsilon$-Greedy, UCB, and TS respectively. We consider four different cases: two cases for the abruptly changing scenario, and two for the drifting scenario; each case has 100 independent replications. The length of each simulated experiment is $T=10,000$. 

\subsubsection{Abruptly Changing Expected Reward}
\label{sec-AbruptChange}

\begin{figure}[H]
    \centering
    \subfloat[]
    {\includegraphics[width=0.33\textwidth, height=0.16\textheight]{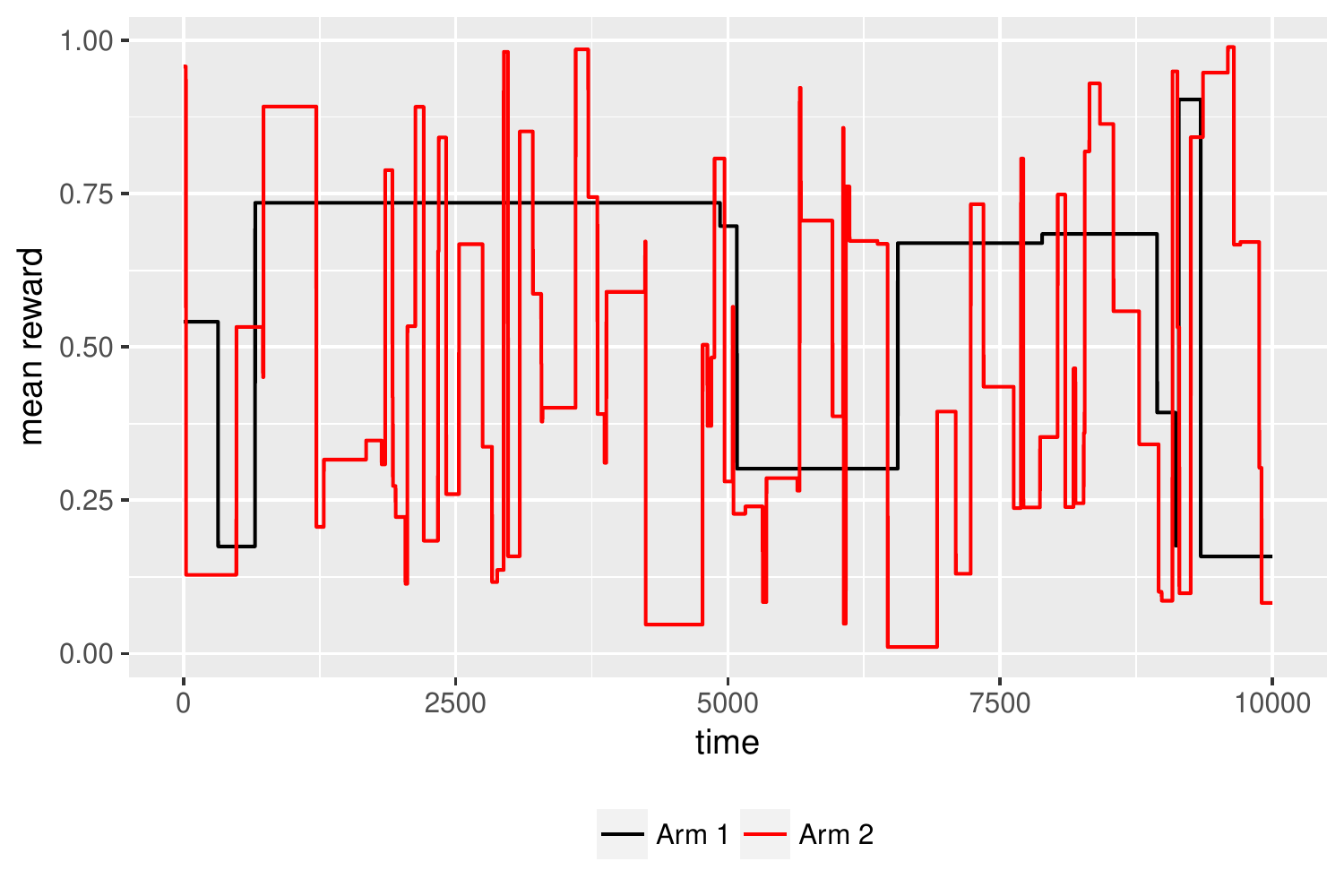}}\hfill
    \subfloat[]
    {\includegraphics[width=0.33\textwidth, height=0.16\textheight]{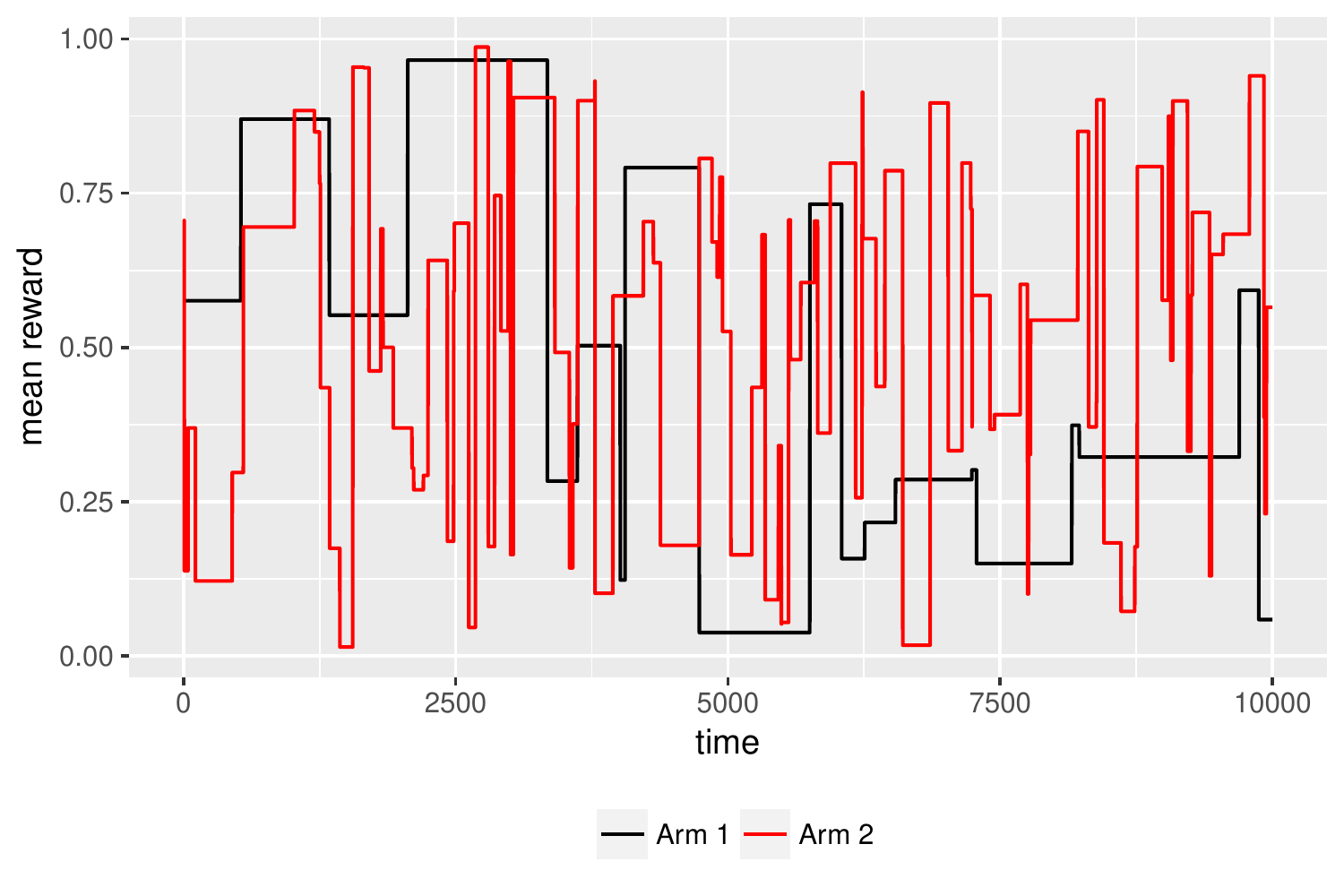}}\hfill
    \subfloat[]
    {\includegraphics[width=0.33\textwidth, height=0.16\textheight]{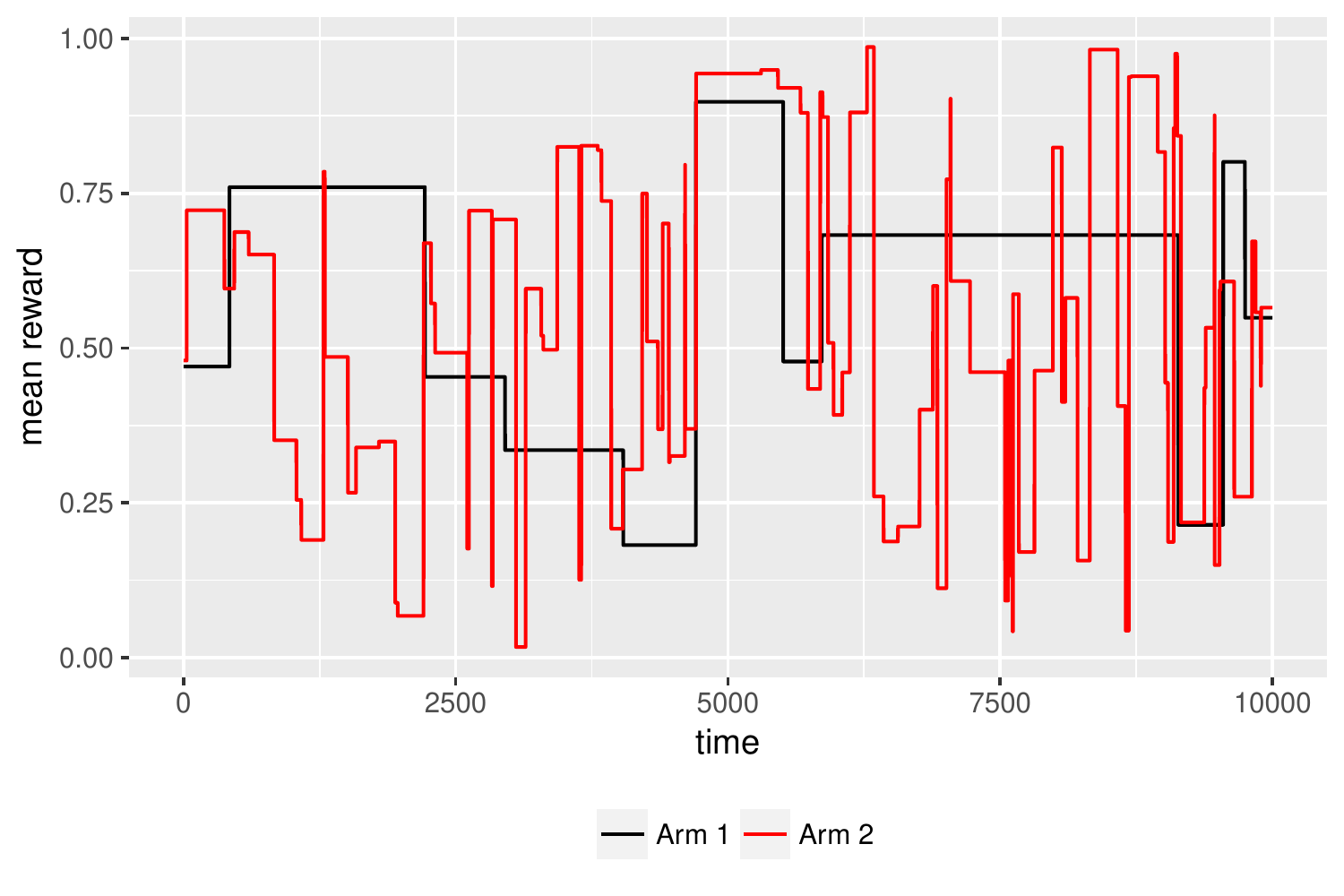}}\hfill
    
    \caption{Abruptly changing scenario (Case~1): examples $\mu_{t}$ sampled from the model in (\ref{eqt-AbruptChangeModel}) with parameters of Case~1 displayed in Table~\ref{table-ParaD1D2}.}
    \label{plot-sample-Uniform-D1} 
\end{figure}

\begin{figure}[H]
    \centering
    \subfloat[]
    {\includegraphics[width=0.33\textwidth, height=0.16\textheight]{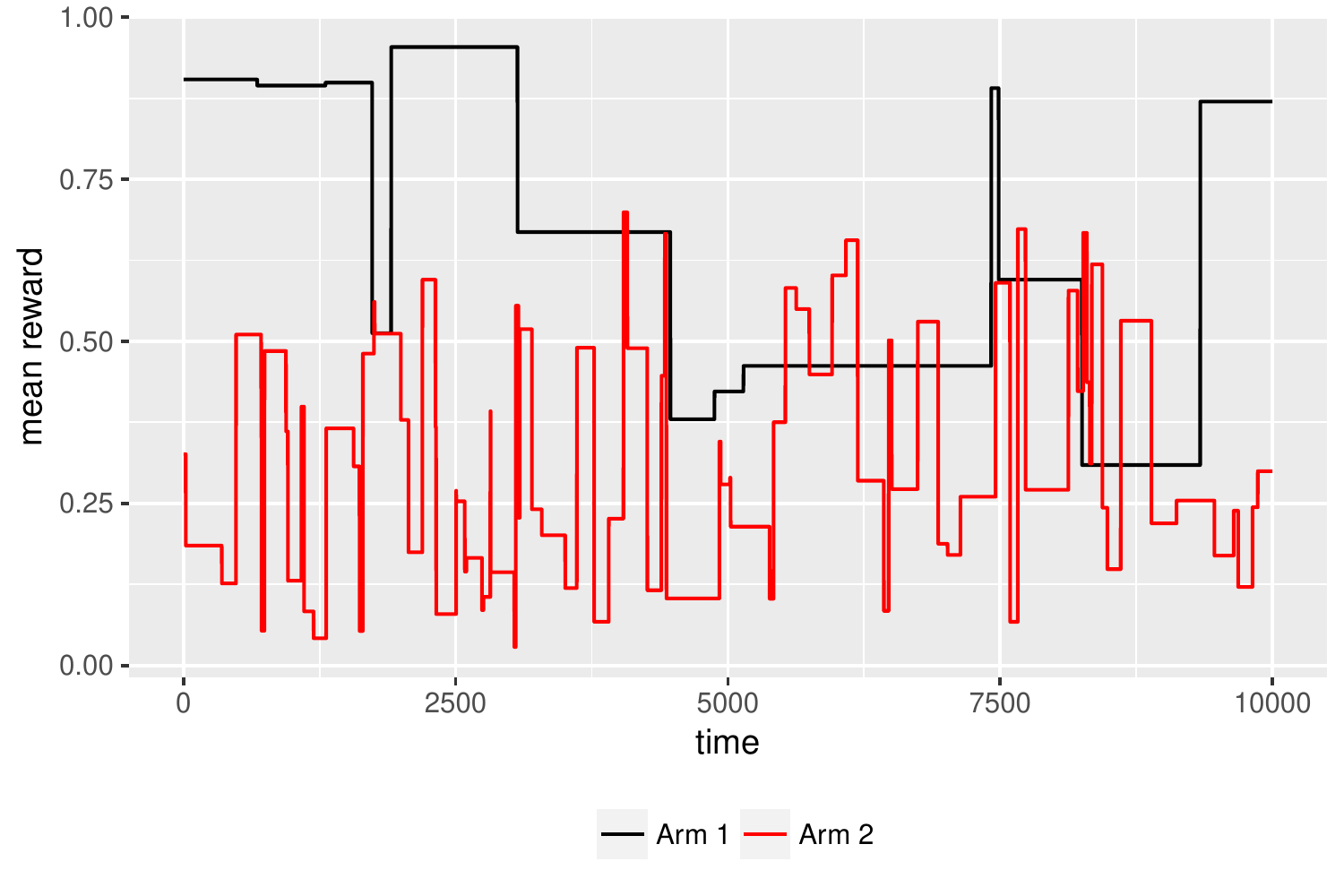}}\hfill
    \subfloat[]
    {\includegraphics[width=0.33\textwidth, height=0.16\textheight]{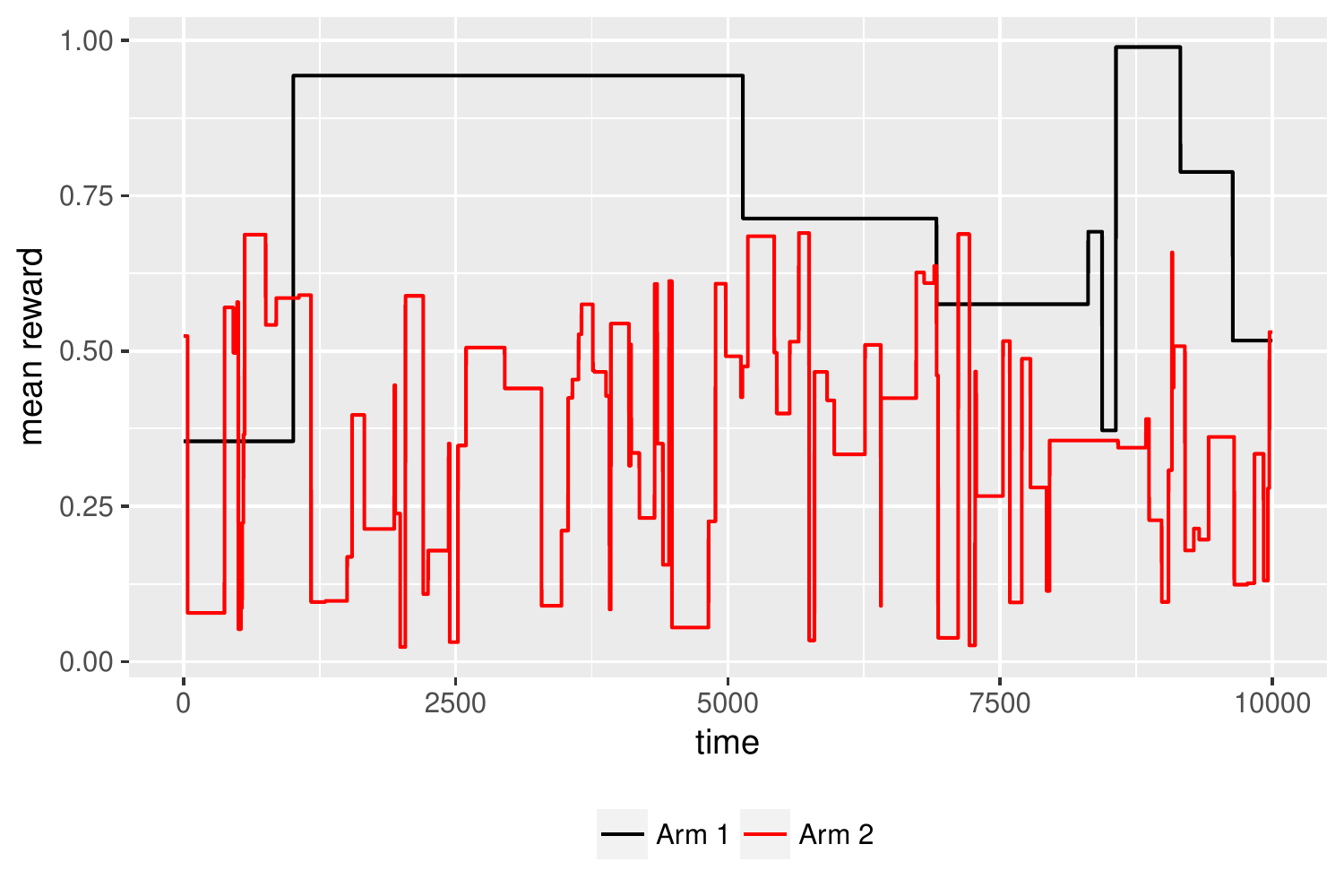}}\hfill
    \subfloat[]
    {\includegraphics[width=0.33\textwidth, height=0.16\textheight]{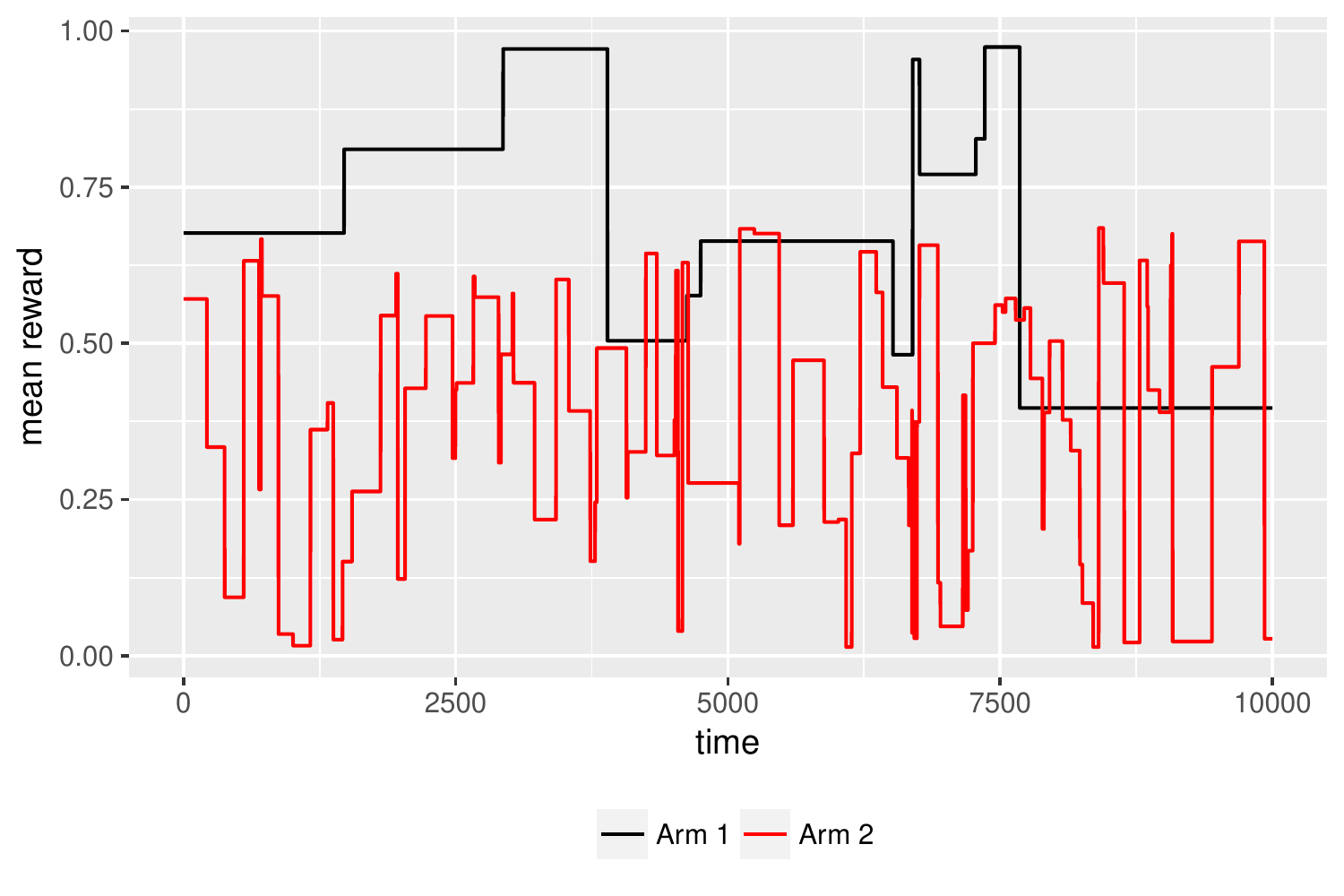}}\hfill
    
    \caption{Abruptly changing scenario (Case~2): examples $\mu_{t}$ sampled from the model in (\ref{eqt-AbruptChangeModel}) with parameters of Case~2 displayed in Table~\ref{table-ParaD1D2}.}
    \label{plot-sample-Uniform-D3} 
\end{figure}

The expected reward $\mu_{t}$ is simulated by the following exponential clock model:
\begin{align}
\label{eqt-AbruptChangeModel}
 \left \{ 
\begin{array}{l}
  J(t)  \sim \text{homogeneous Poisson process with rate} ~ \theta, \\
 \mu_{t}  = \left \{ 
\begin{array}{ll}
\text{sample from~} \mathcal{U} (r_{l}, r_{u}), & \text{if} ~~ J(t) > J(t-1), \\
\mu_{t-1}, & \text{otherwise},
\end{array}
\right. \\
 \mu_{0}  = \text{sample from~} \mathcal{U} (r_{l}, r_{u}).
 \end{array}
\right.
\end{align}
The parameter $\theta$ determines the frequency at which change point occurs. At each change point,  the new expected reward is sampled from a uniform distribution $\mathcal{U} (r_{l}, r_{u})$. We generate two different cases, Case~1~and~2. Parameters used for generating these cases can be found in Table~\ref{table-ParaD1D2}. For visualisation purposes, we display three examples of simulated path $\mu_t$ for Case~1/2 in Figure~\ref{plot-sample-Uniform-D1}/\ref{plot-sample-Uniform-D3} respectively. For Case~1, we distinguish the two arms by varying their frequency of change, but in the long run, for high $T$, $\bar{\mu}_{T} = \mathbb{E}[\frac{1}{T} \sum_{i=1}^{T} \mu_{i}]$ are the same. In Case~2, Arm~1 has a higher $\bar{\mu}_{T}$.

\begin{table}[H]
\centering
\caption{Parameters used in the exponential clock model shown in (\ref{eqt-AbruptChangeModel}).}     
    \label{table-ParaD1D2}
  \begin{tabular}{c | c c c | c c c  }
     \hline
    \multirow{2}{*}{} &
      \multicolumn{3}{c |}{Case 1} &
      \multicolumn{3}{c}{Case 2} \\
    & $\theta$ &  $r_{l}$ & $r_{u}$ & $\theta$ &  $r_{l}$ & $r_{u}$ \\
    \hline
    Arm 1 & 0.001 & 0.0 & 1.0 & 0.001 & 0.3 & 1.0 \\
    
    Arm 2 & 0.010  & 0.0 & 1.0 & 0.010 & 0.0 & 0.7 \\
    \hline
  \end{tabular}
\end{table}

\begin{figure}[H]
    \centering
    \subfloat[Case 1: cumulative regret.]
    {\includegraphics[width=0.45\textwidth, height=0.2\textheight]{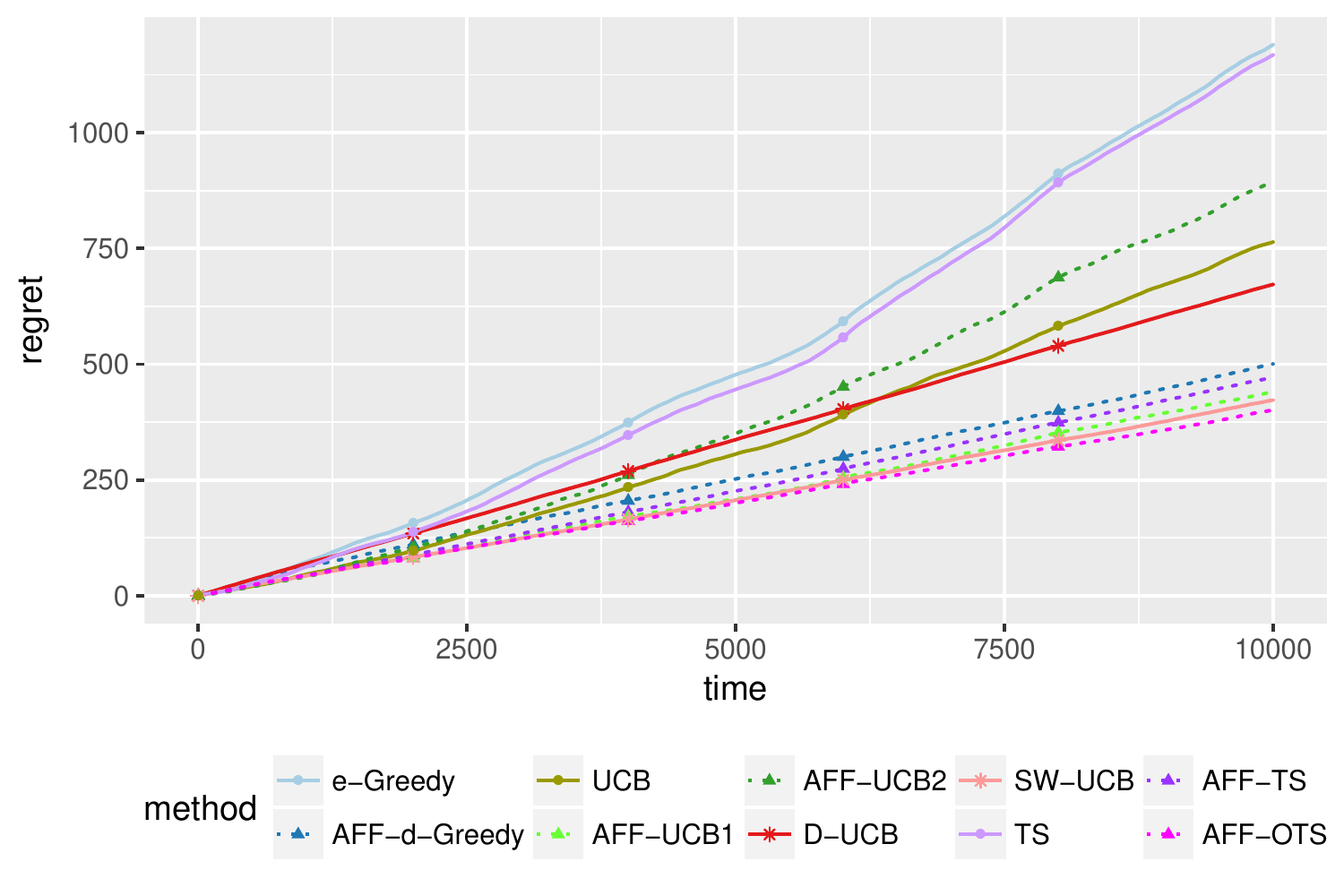}}\hfill
    \subfloat[Case 2: cumulative regret.]
    {\includegraphics[width=0.45\textwidth, height=0.2\textheight]{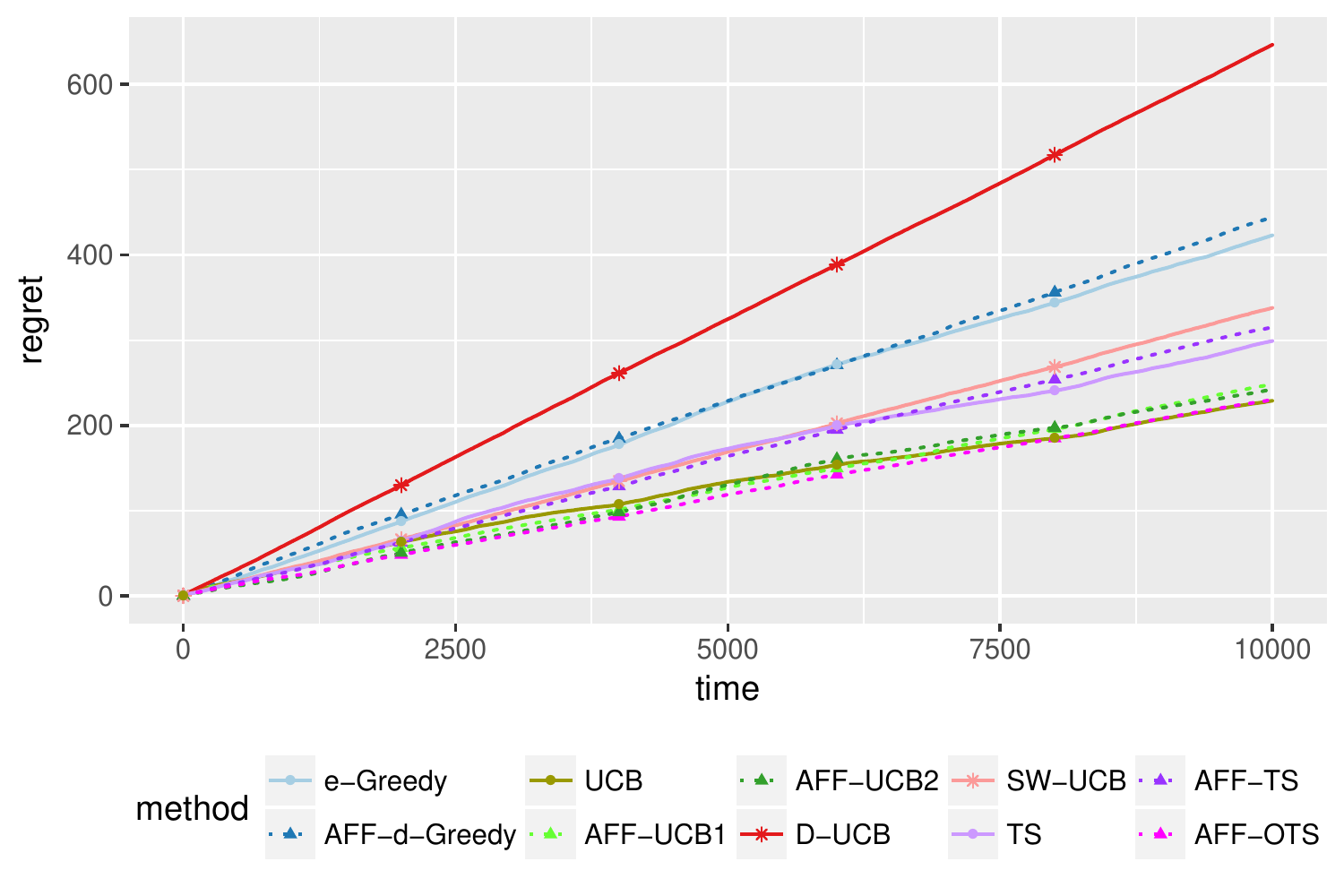}}\hfill
    \subfloat[Case 1: boxplot of total regret.]
    {\includegraphics[width=0.45\textwidth, height=0.2\textheight]{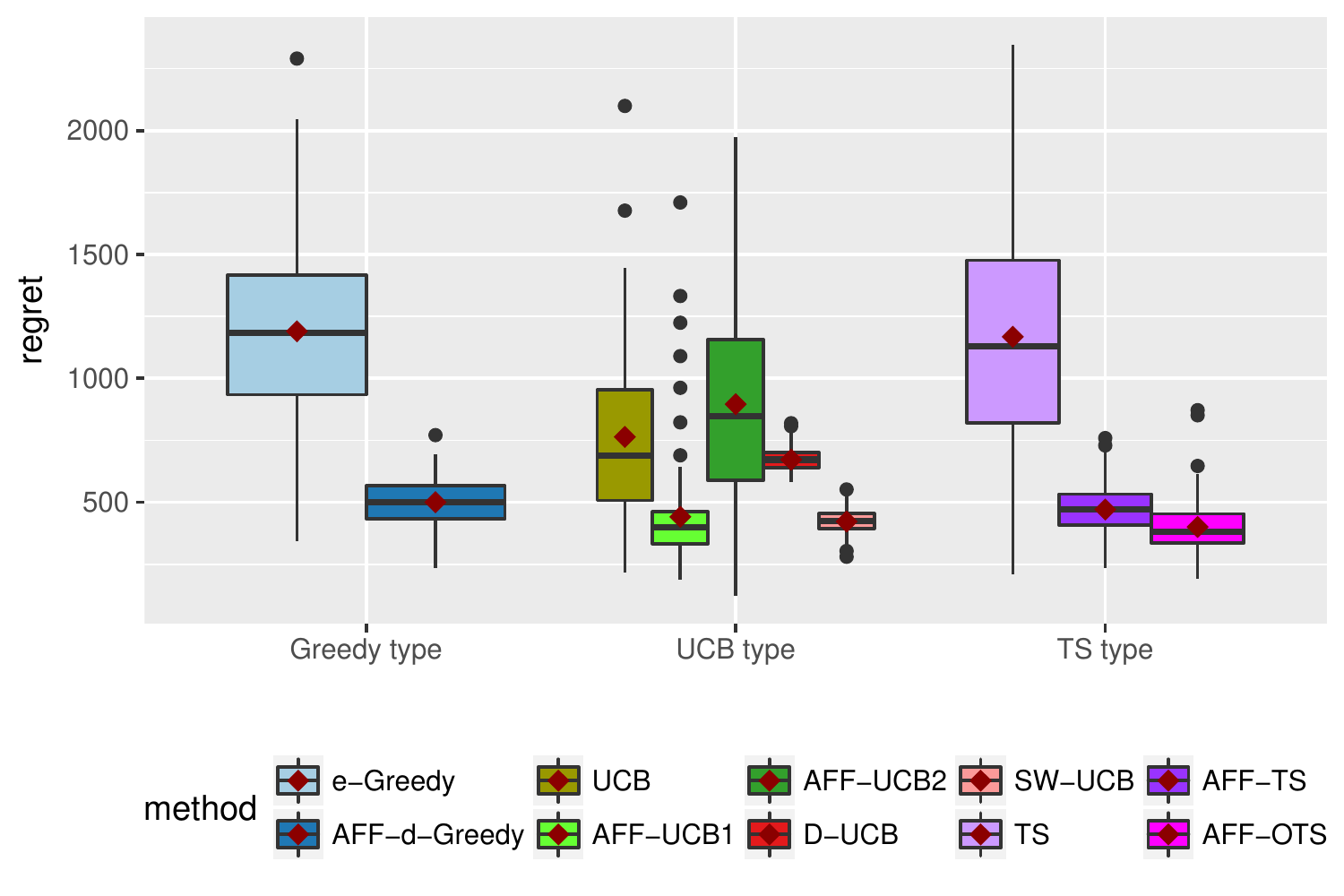}}\hfill
    \subfloat[Case 2: boxplot of total regret.]
    {\includegraphics[width=0.45\textwidth, height=0.2\textheight]{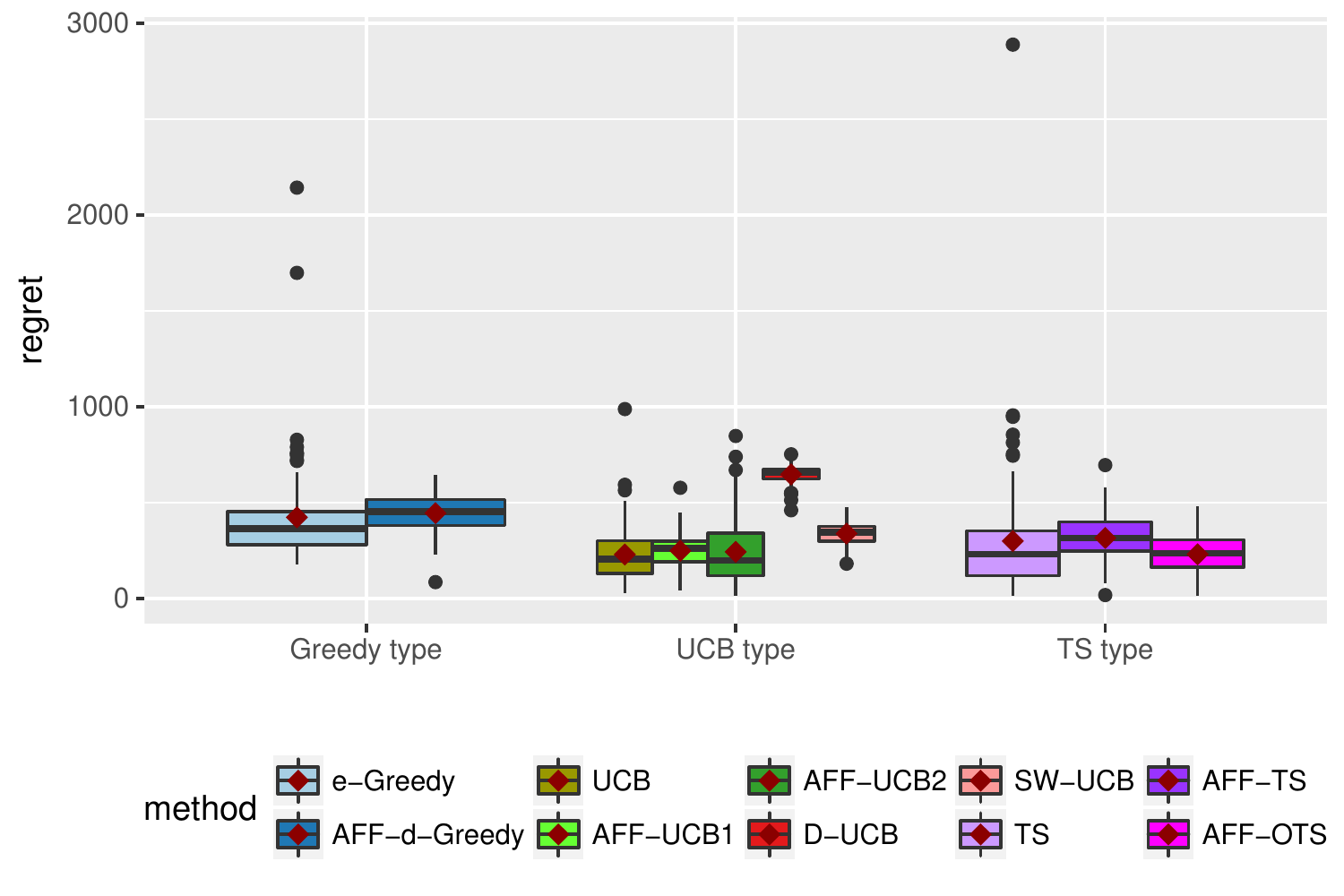}}\hfill
                    
    \caption{Results for the two-armed Bernoulli bandit with abruptly changing expected rewards. The top row displays the cumulative regret over time; results are averaged over 100 replications. The bottom row are boxplots of total regret at time $t=10,000$. Trajectories are sampled from (\ref{eqt-AbruptChangeModel}) with parameters displayed in Table~\ref{table-ParaD1D2}.}
    \label{plot-D1D2} 
\end{figure}

In Figure~\ref{plot-D1D2}, we present comparisons in each abruptly changing case. The bottom row of Figure~\ref{plot-D1D2} shows boxplots of the total regret $R_{T}$ as in (\ref{seq-regret}). In addition, the top row of Figure~\ref{plot-D1D2} displays the cumulative regret over time; the results are averaged over 100 independent replications. The plots are good evidence that our algorithms yield improved performance over standard approaches. In particular, the improvement is distinguishable in Case~1, for which the two arms have the same $\bar{\mu}_{T}$.   
Our AFF-deployed UCB methods perform differently: AFF-UCB1 improves the standard UCB method apparently while AFF-UCB2 performs slightly worse than UCB. Furthermore, AFF-$d$-Greedy, AFF-UCB1, AFF-TS, AFF-OTS, and SW-UCB have similar performance. In the case that one arm's mean dominates in the long run (Case~2), the AFF MAB algorithms perform similarly to standard methods. However, the AFF MAB algorithms (except AFF-UCB2) have smaller variance among replications. In both cases, AFF-OTS has the best (or nearly best) performance in terms of total regret.


%
%

%


\subsubsection{Drifting Expected Reward}
\label{sec-Drifting}

\begin{figure}[H]
    \centering
    \subfloat[]
    {\includegraphics[width=0.33\textwidth, height=0.16\textheight]{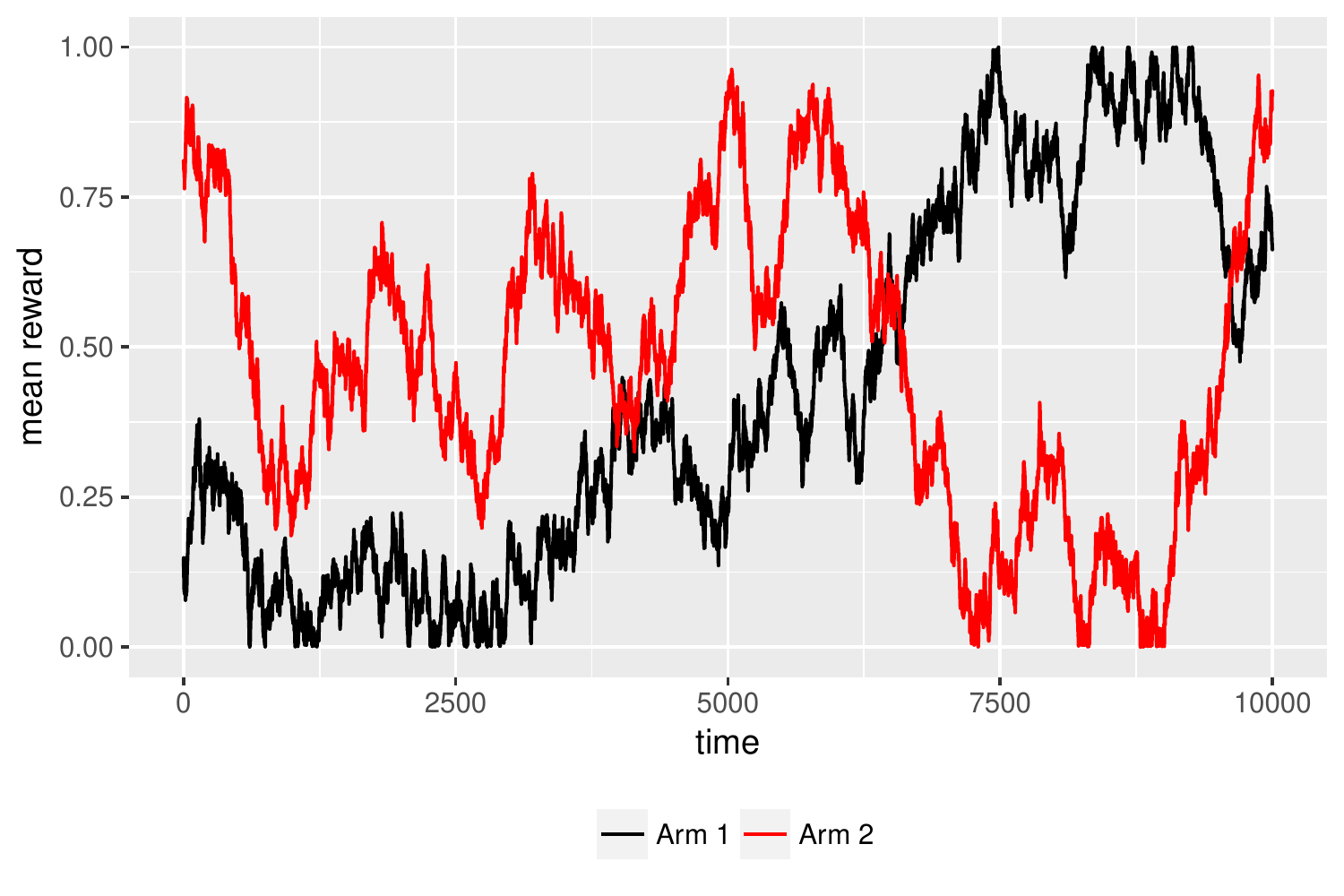}}\hfill
    \subfloat[]
    {\includegraphics[width=0.33\textwidth, height=0.16\textheight]{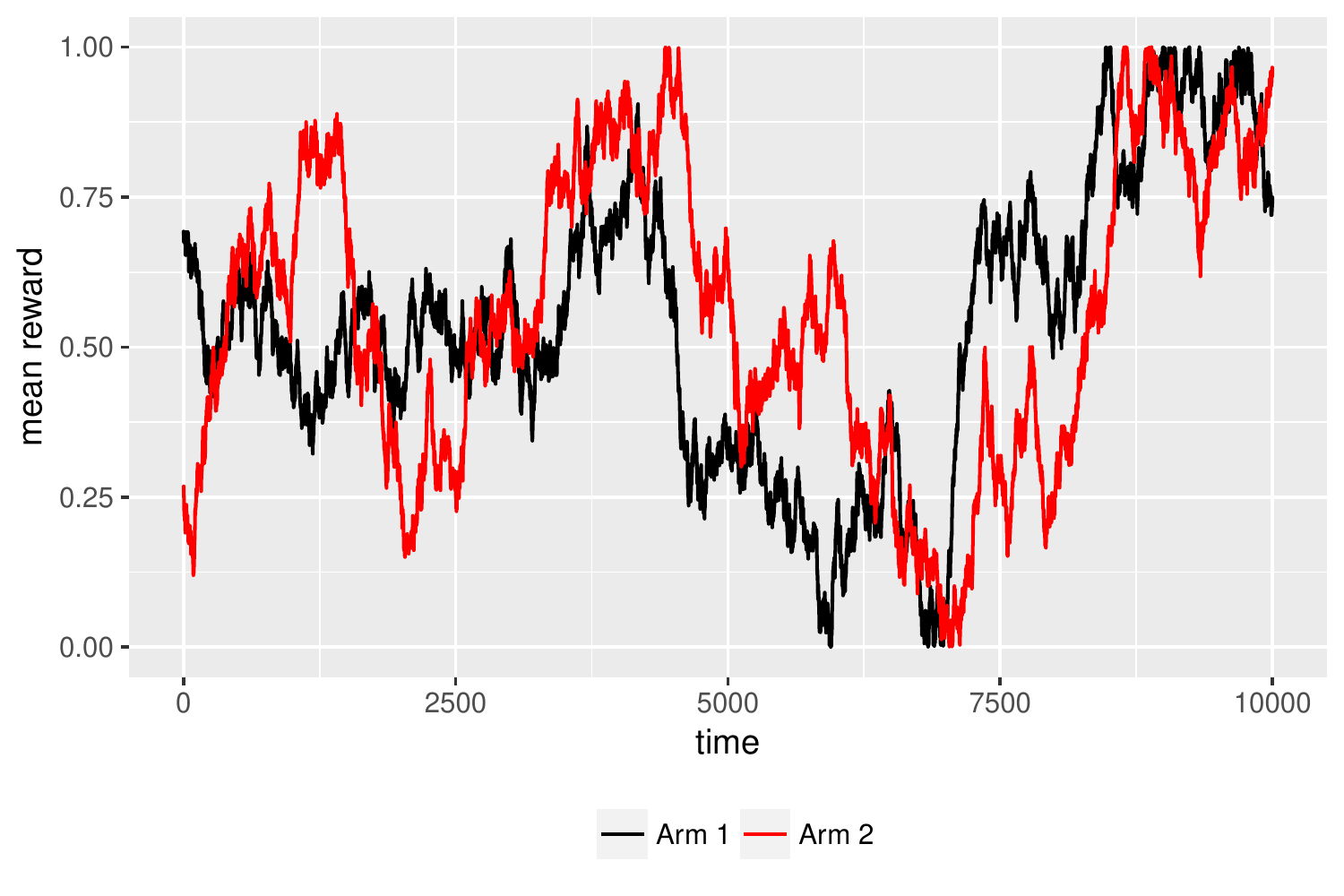}}\hfill
    \subfloat[]
    {\includegraphics[width=0.33\textwidth, height=0.16\textheight]{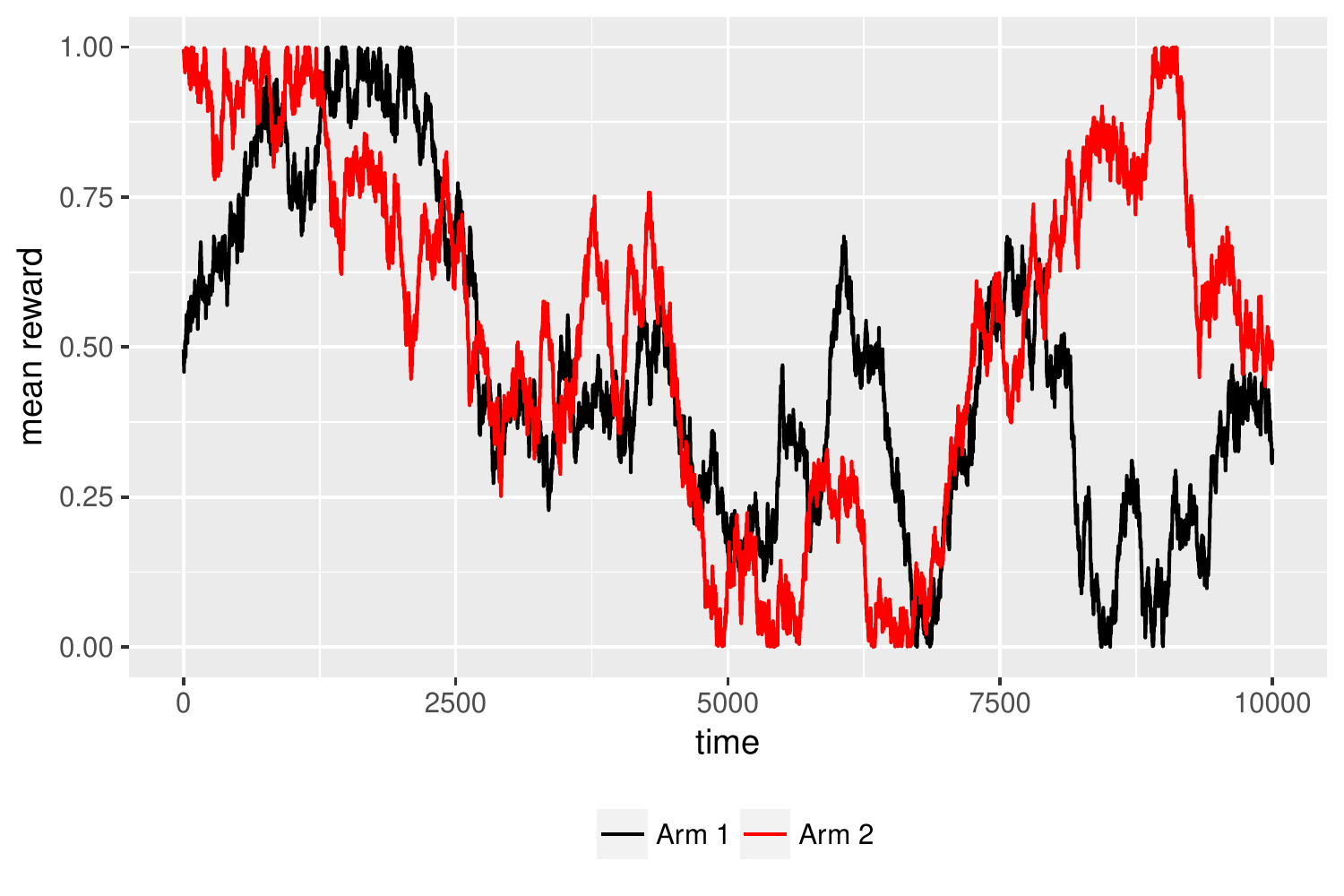}}\hfill
    
    \caption{Drifting scenario (Case~3): examples of simulated $\mu_t$ from the model in (\ref{eqt-DriftingModel}) with $\sigma^{2}_{\mu}=0.0001$.}
    \label{plot-sample-RWRB-D1} 
\end{figure}

For the drifting scenario, we use two different models. The first is the random walk model with reflecting bounds introduced in \citet{Slivkins2008}, which is:
\begin{align}
\label{eqt-DriftingModel}
& \mu_{t} = f(\mu_{t-1} + \omega_{t}), ~~ \omega_{t} \sim N(0, \sigma^{2}_{\mu}) ,
\end{align}
where $f(x)=\left \{ \begin{array}{l  l}
x' & x' \leq 1 \\
1- (x'-1)  & x' > 1
\end{array}
\right.$, and $x' = |x| ~(\text{mod} ~2)$.
\citet{Slivkins2008} showed that $\mu_{t}$ generated by this model is stationary, that is in the long run, $\mu_{t}$ will be distributed according to a uniform distribution. The parameter $\sigma^{2}_{\mu}$ used in the model controls the rate of change in an arm. In Figure~\ref{plot-sample-RWRB-D1}, we illustrate three sample paths $\mu_{t}$ against $t$ simulated via (\ref{eqt-DriftingModel}) with $\sigma^{2}_{\mu}=0.0001$ (Case~3). Similar to Case~1, the two arms in Case~3 have the same $\bar{\mu}_{T}$.

\begin{figure}[H]
    \centering
    \subfloat[]
    {\includegraphics[width=0.33\textwidth, height=0.16\textheight]{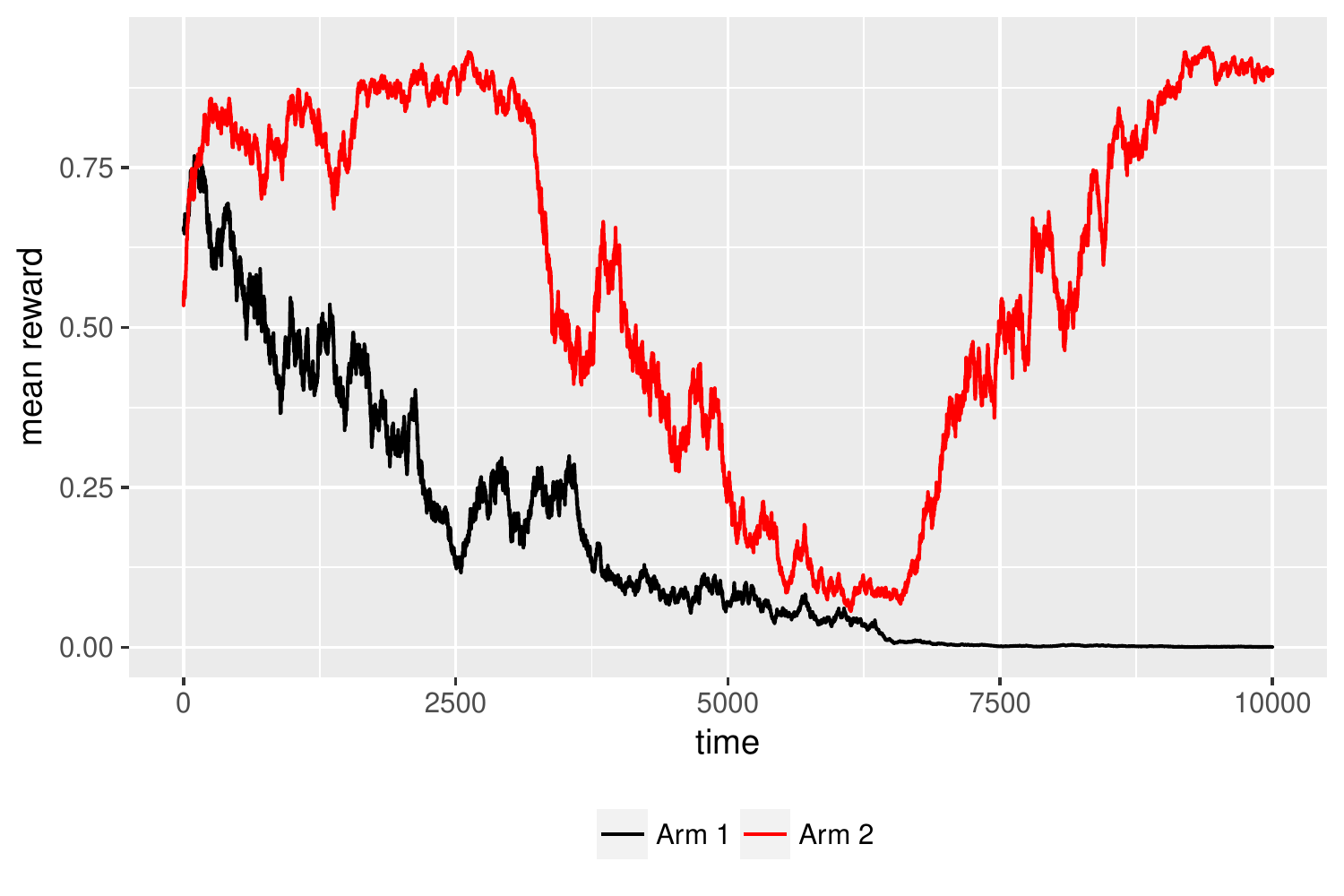}}\hfill
    \subfloat[]
    {\includegraphics[width=0.33\textwidth, height=0.16\textheight]{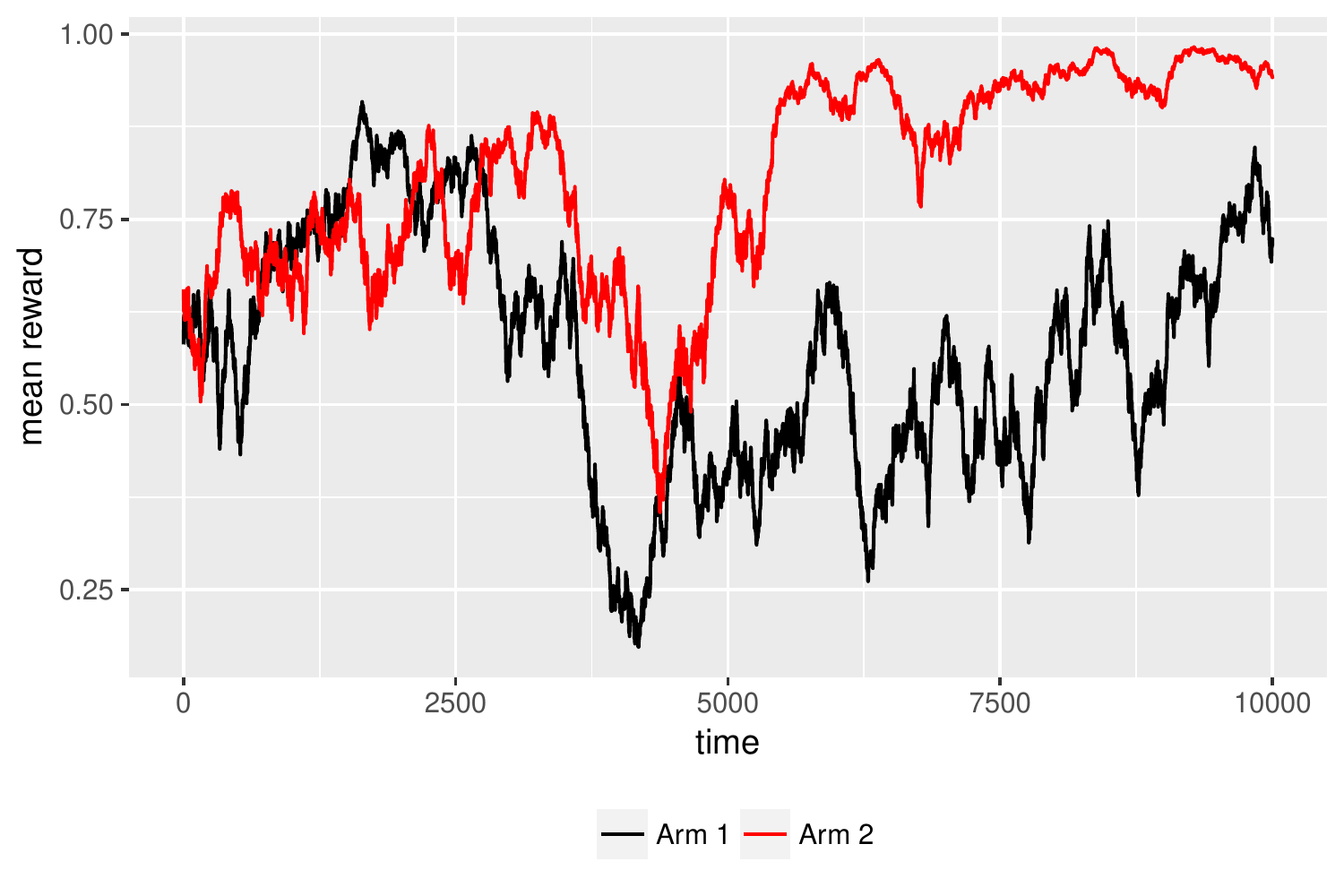}}\hfill
    \subfloat[]
    {\includegraphics[width=0.33\textwidth, height=0.16\textheight]{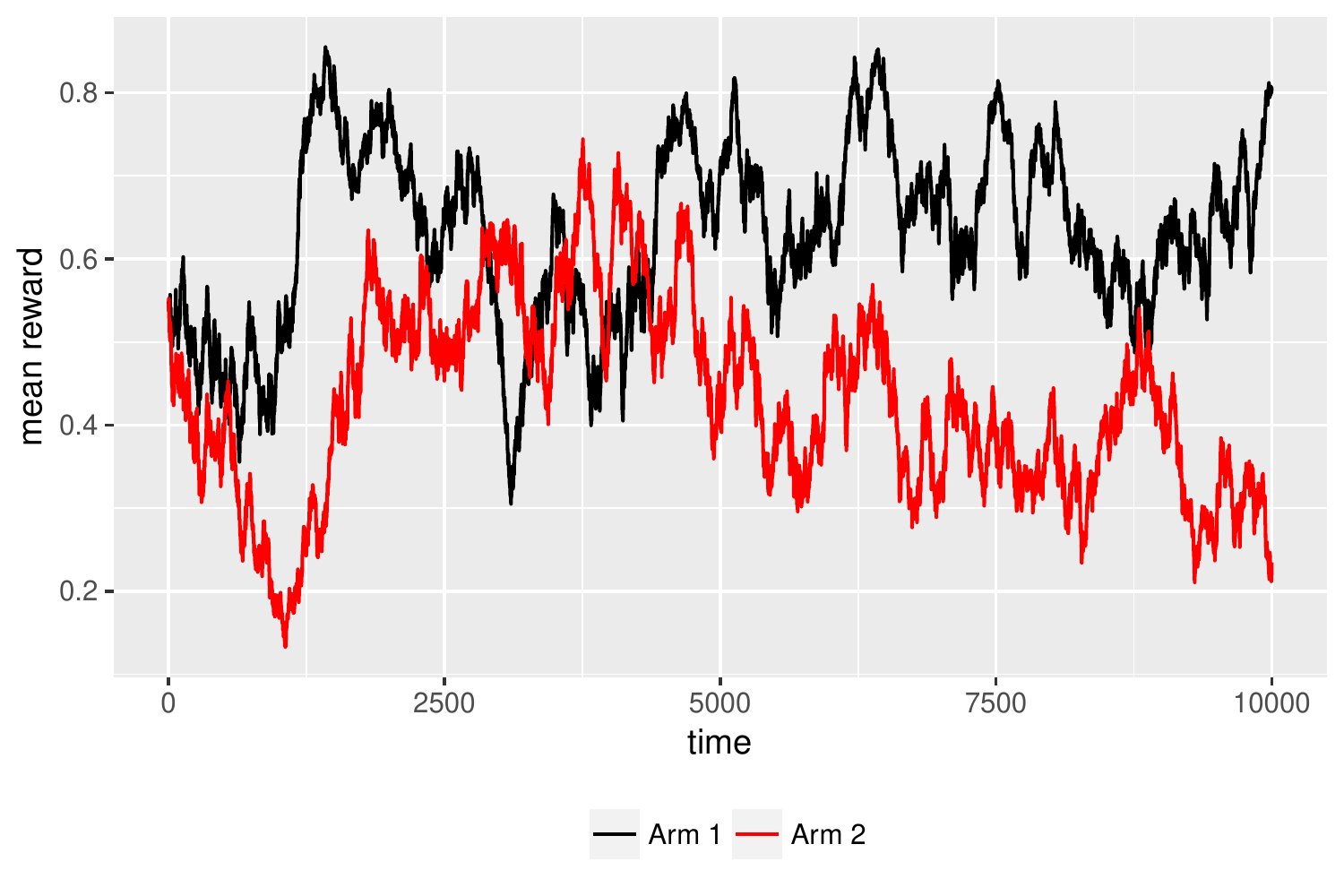}}\hfill
    
    \caption{Drifting scenario (Case~4): examples of simulated $\mu_t$ from the model in (\ref{eqt-DriftingModel2}) with $\sigma^{2}_{\mu}=0.001$.}
    \label{plot-sample-RWLO-D2} 
\end{figure}

The second model we use to simulate drifting arms is:
\begin{align}
\label{eqt-DriftingModel2}
 \left \{ 
\begin{array}{l}
z_{0}  = \text{sample from~} \mathcal{U} (0, 1), \\
  z_{t} = z_{t-1} +  \omega_{t}, ~~ \omega_{t} \sim N(0, \sigma^{2}_{\mu}), \\
  \mu_{t} = \frac{1}{1+\exp(-z_{t})},
 \end{array}
\right.
\end{align}
where the expected reward $\mu_{t}$ is transformed from the random walk $z_{t}$; the parameter $\sigma^{2}_{\mu}$ controls the speed that $\mu_{t}$ evolves. Since a random walk diverges in the long run, any trajectory will move closer and closer to one of the boundaries 0 or 1; the two arms generated from this model can either move toward the same boundary or separate in the long run. Figure~\ref{plot-sample-RWLO-D2} displays three sample paths generated via (\ref{eqt-DriftingModel2}) with $\sigma^{2}_{\mu}=0.001$ (Case~4).

\begin{figure}[H]
    \centering
    \subfloat[Case 3: cumulative regret.]
    {\includegraphics[width=0.45\textwidth, height=0.2\textheight]{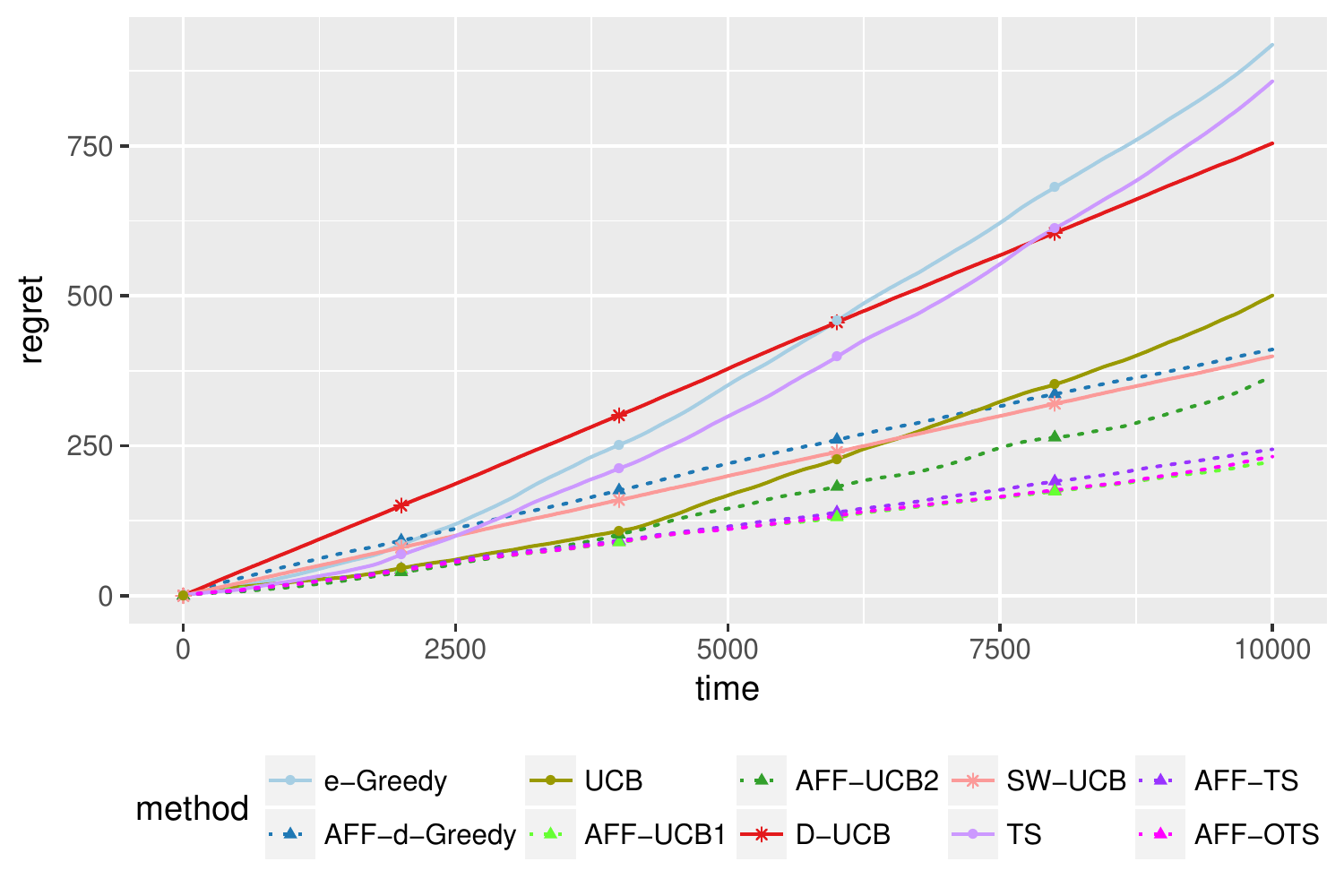}}\hfill
    \subfloat[Case 4: cumulative regret.]
    {\includegraphics[width=0.45\textwidth, height=0.2\textheight]{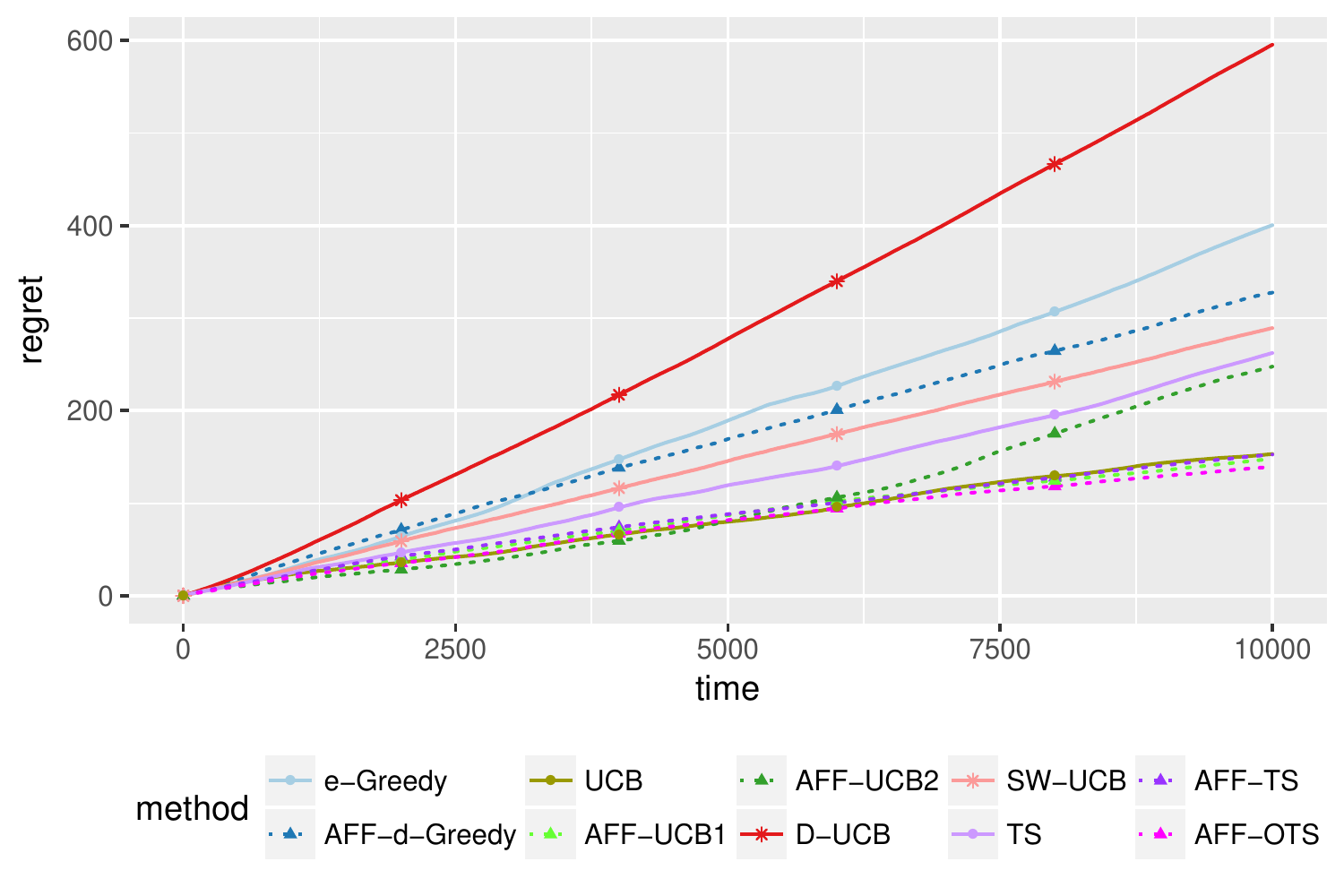}}\hfill
    \subfloat[Case 3: boxplot of total regret.]
    {\includegraphics[width=0.45\textwidth, height=0.2\textheight]{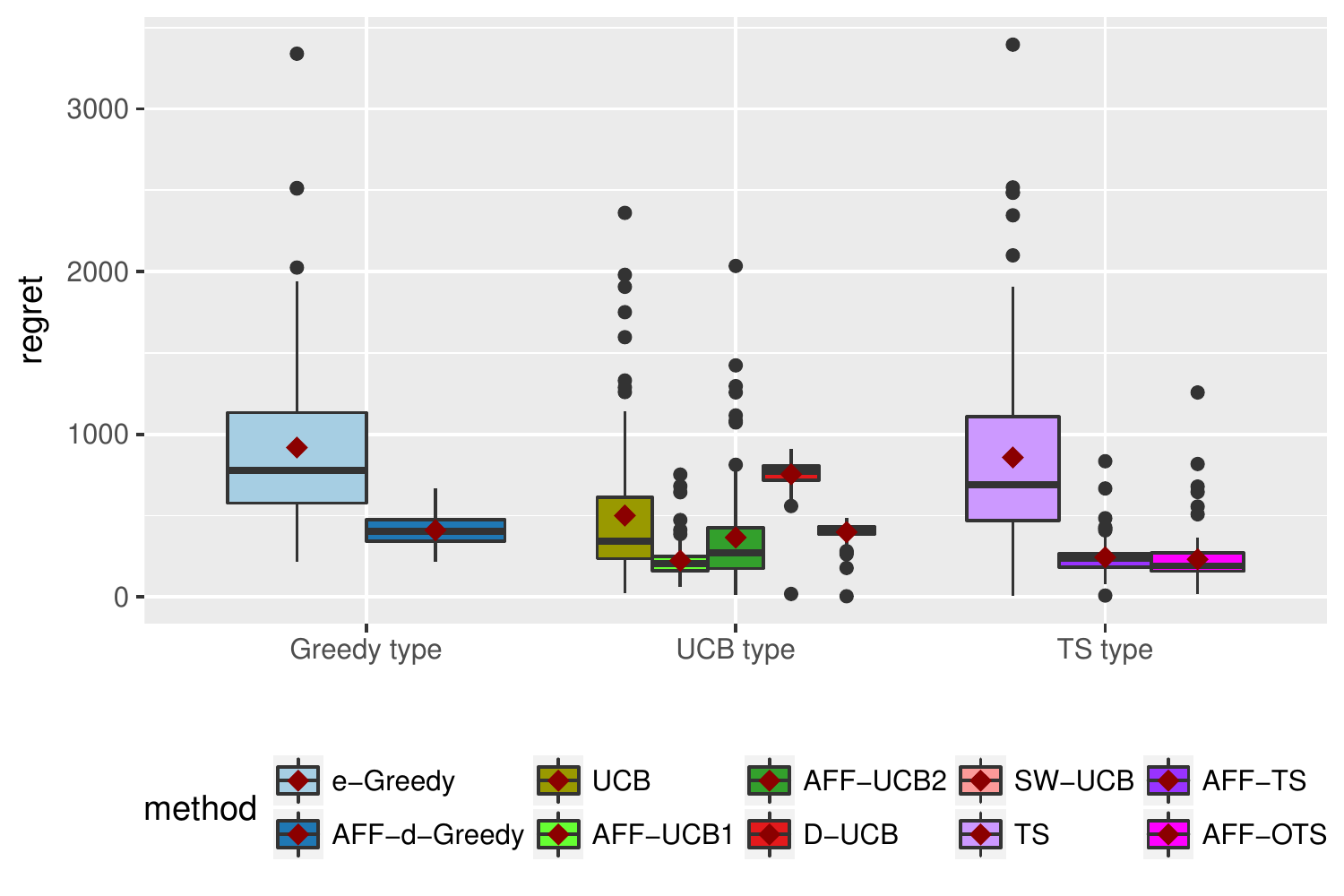}}\hfill
    \subfloat[Case 4: boxplot of total regret.]
    {\includegraphics[width=0.45\textwidth, height=0.2\textheight]{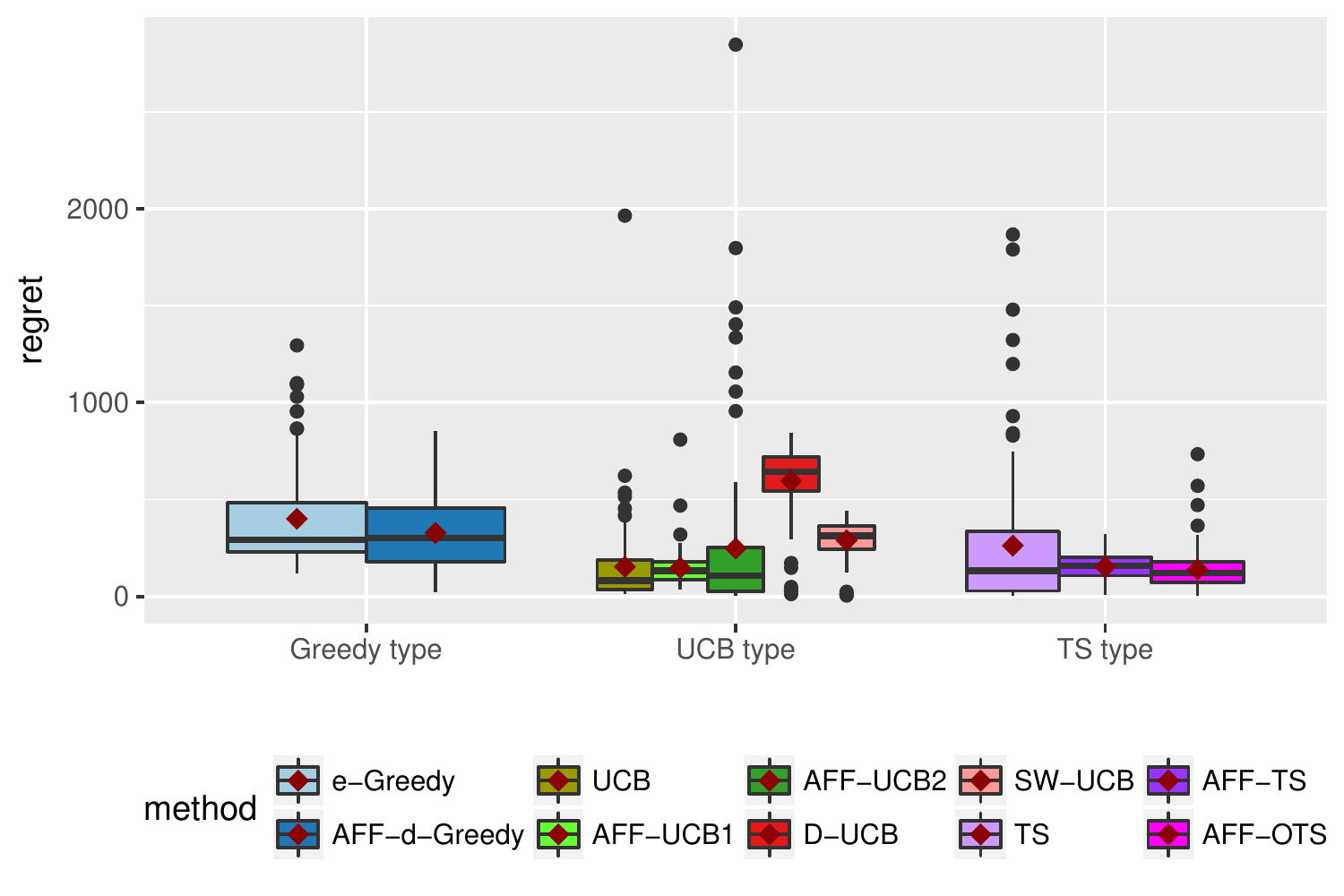}}\hfill
                     
    \caption{Results for the two-armed Bernoulli bandit with drifting expected rewards. The top row displays the cumulative regret over time; results are averaged over 100 independent replications. The bottom row are boxplots of total regret at time $t=10,000$. Trajectories for Case~3 are sampled from (\ref{eqt-DriftingModel}) with $\sigma^{2}_{\mu} = 0.0001$, and trajectories for Case~4 are sampled from (\ref{eqt-DriftingModel2}) with $\sigma^{2}_{\mu} = 0.001$.}
    \label{plot-D3D4} 
\end{figure}

The results for the drifting scenario can be found in Figure~\ref{plot-D3D4}. The top row of Figure~\ref{plot-D3D4} displays the cumulative regret averaged over 100 independent replications, and the bottom row shows boxplots of total regret. For Case~3 that is simulated from the model in (\ref{eqt-DriftingModel}), we can see that the AFF MAB algorithms outperform standard approaches. For Case~4 that is simulated from the model in (\ref{eqt-DriftingModel2}), there is a solid improvement in the performance of TS, while UCB and AFF-UCB1/AFF-UCB2 perform similarly. Similar to the abruptly changing case, AFF-OTS performs very well in both drifting cases in terms of total regret. It was more challenging to deploy adaptive estimation with UCB because it was harder to interpret the estimate from AFF estimator (it is more dynamic with less memory) and modify the upper bound.

\subsubsection{Large Number of Arms}
\begin{figure}[H]
    \centering
         \includegraphics[width=0.9\textwidth, height=0.25\textheight]{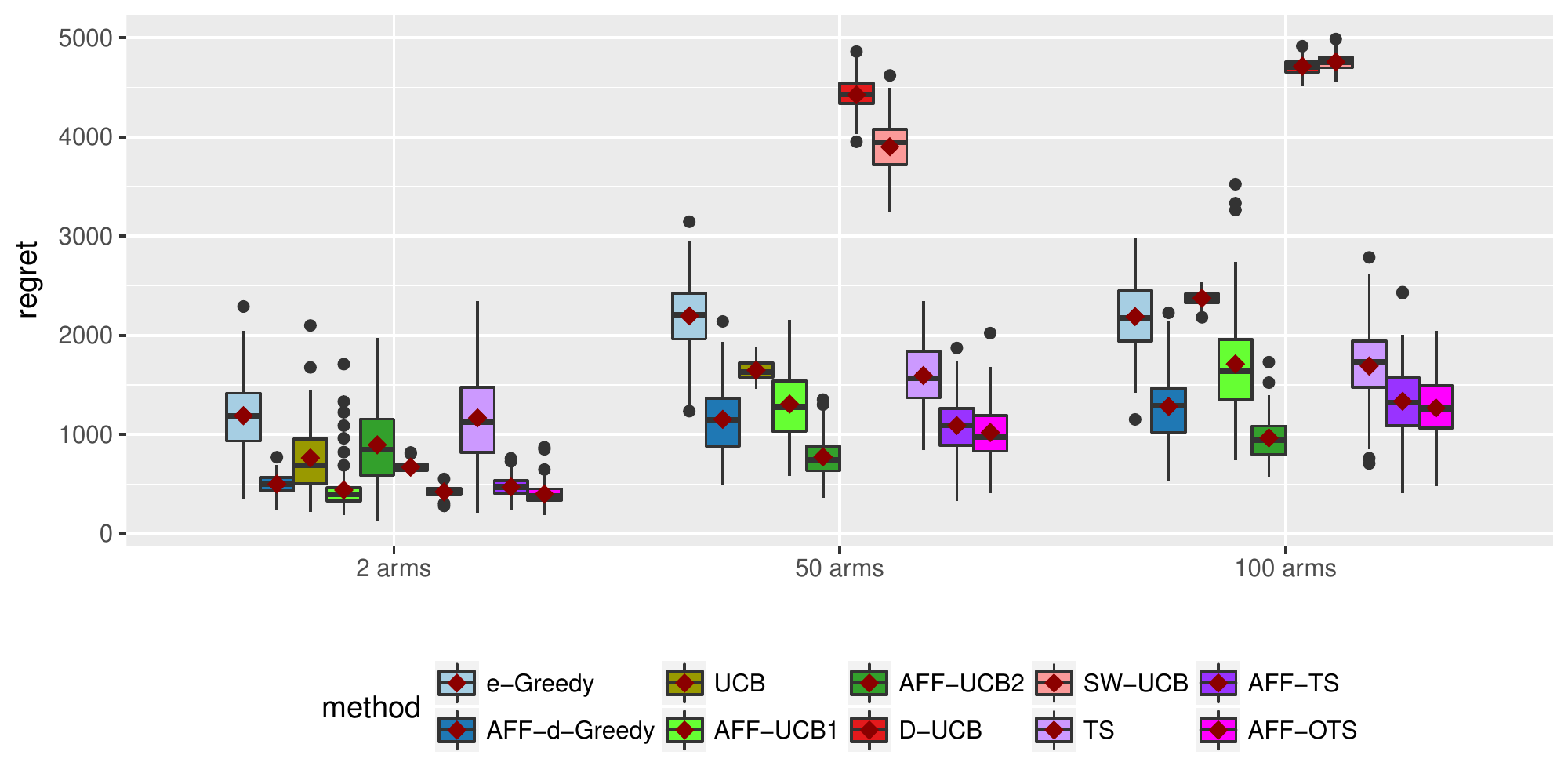}
    \caption{Large number of arms: abruptly changing environment (Case~1).}
    \label{plot-Uniform-multi} 
\end{figure}

\begin{figure}[H]
    \centering
         \includegraphics[width=0.9\textwidth, height=0.25\textheight]{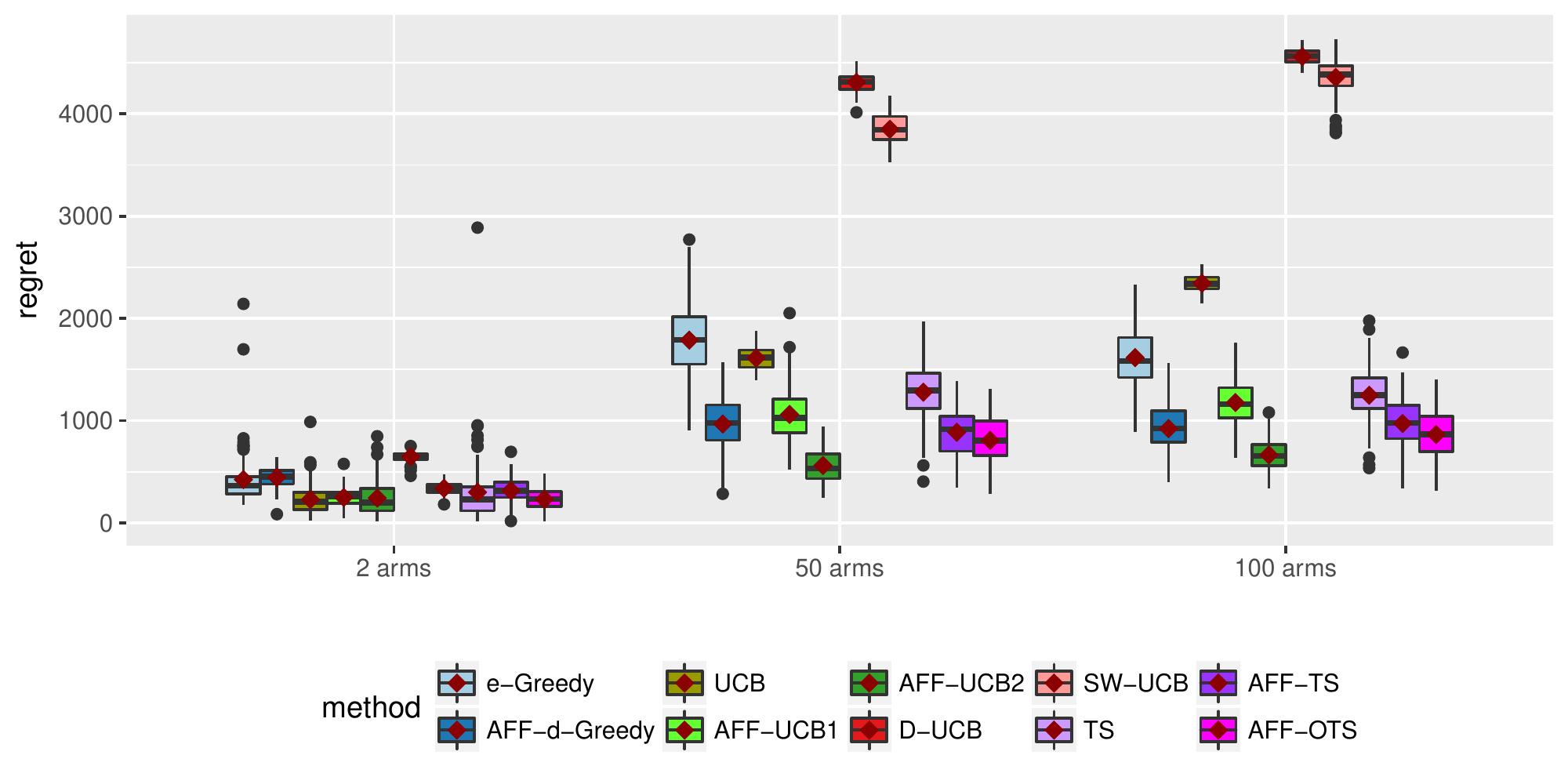}
    \caption{Large number of arms: abruptly changing environment (Case~2).}
    \label{plot-Uniform-multi2} 
\end{figure}

\begin{figure}[H]
    \centering
         \includegraphics[width=0.9\textwidth, height=0.25\textheight]{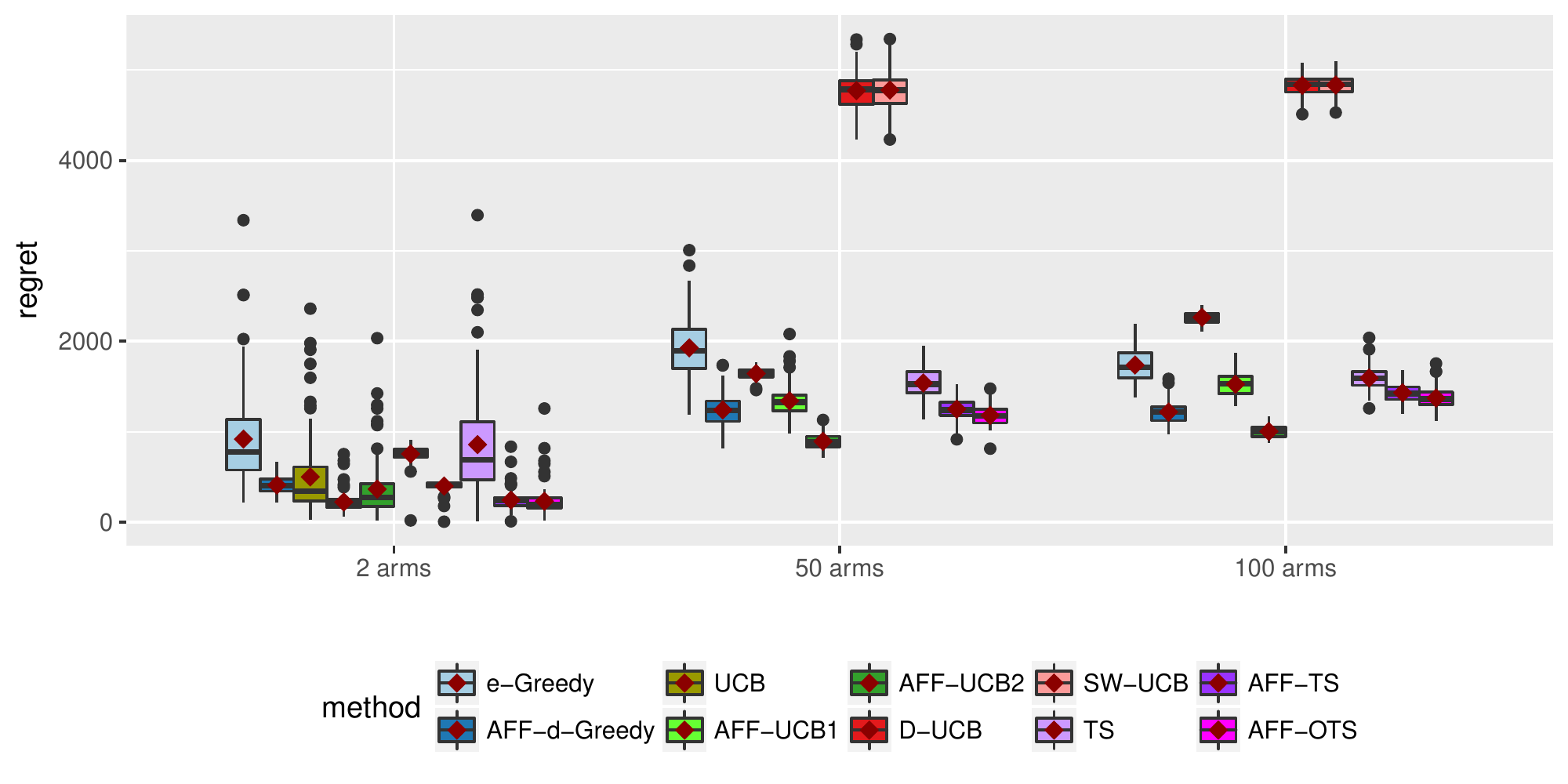}
    \caption{Large number of arms: drifting environment (Case~3).}
    \label{plot-RWRB-multi} 
\end{figure}

\begin{figure}[H]
    \centering
         \includegraphics[width=0.9\textwidth, height=0.25\textheight]{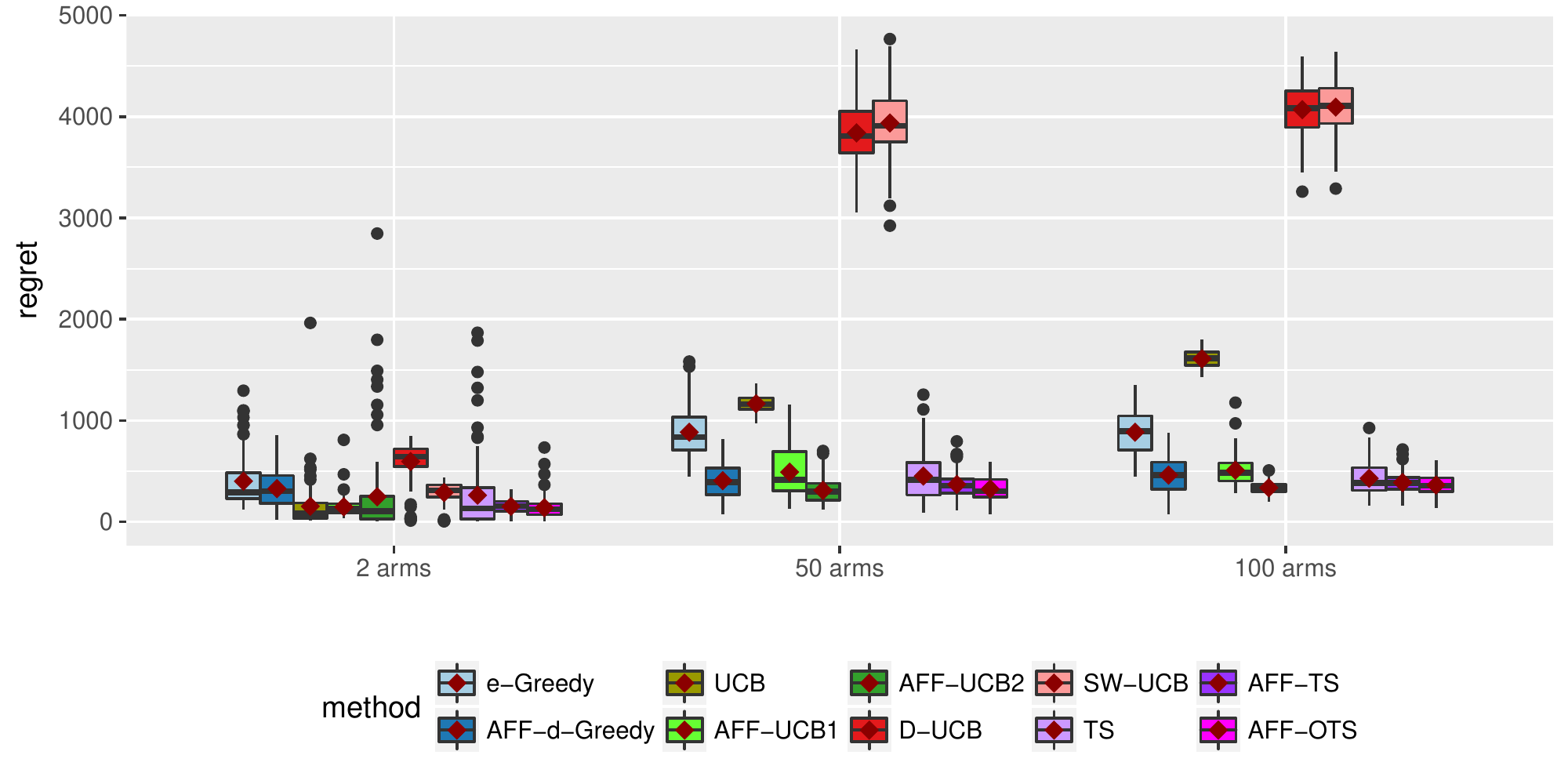}
    \caption{Large number of arms: drifting environment (Case~4).}
    \label{plot-RWLO-multi} 
\end{figure}

Modern applications of bandit problems can involve a large number of arms. For example, in online advertising, we need to optimise among hundreds of websites. Therefore, we evaluate the performance of our AFF~MAB algorithms with a large number of arms. We repeat earlier experiments with 50 and 100 arms. Tuning parameters are initialised in the same way as in two-armed examples (see the beginning of Section~\ref{sec-exps} for details). The results can be seen from Figures~\ref{plot-Uniform-multi}-\ref{plot-RWLO-multi}. It can be seen that performance gains hold for a large number of arms, and are very pronounced for all methods including UCB (that was more challenging to improve). 
Among all cases, unlike the two-armed examples where the improvement of adaptive estimation on UCB is marginal, with 50 and 100 arms, AFF-UCB1 and AFF-UCB2 performs better than UCB. In particular, AFF-UCB2 has the best performance in all cases, and D-UCB and SW-UCB perform much worse than all the other algorithms.
In addition, AFF-OTS has good performance in all cases. 
In summary, with a large number of arms, our algorithms perform much better than standard methods. Interestingly, in all cases, the results for 50 and 100 arms are very similar. This could be attributed to both 50 and 100 arms being numbers large enough to fill the reward space [0,1] well enough so that the decision maker in either case finds high value arms.

\subsection{Robustness to Tuning}
We have already seen the improvements the AFF MAB algorithms can offer in different dynamic scenarios. We now move on to examine the sensitivity of performance to the tuning parameters. 

\subsubsection{Initialisation in the AFF MAB Algorithms}
\label{sec-initialisation}
\begin{figure}[H]
    \centering
         \includegraphics[width=0.9\textwidth, height=0.17\textheight]{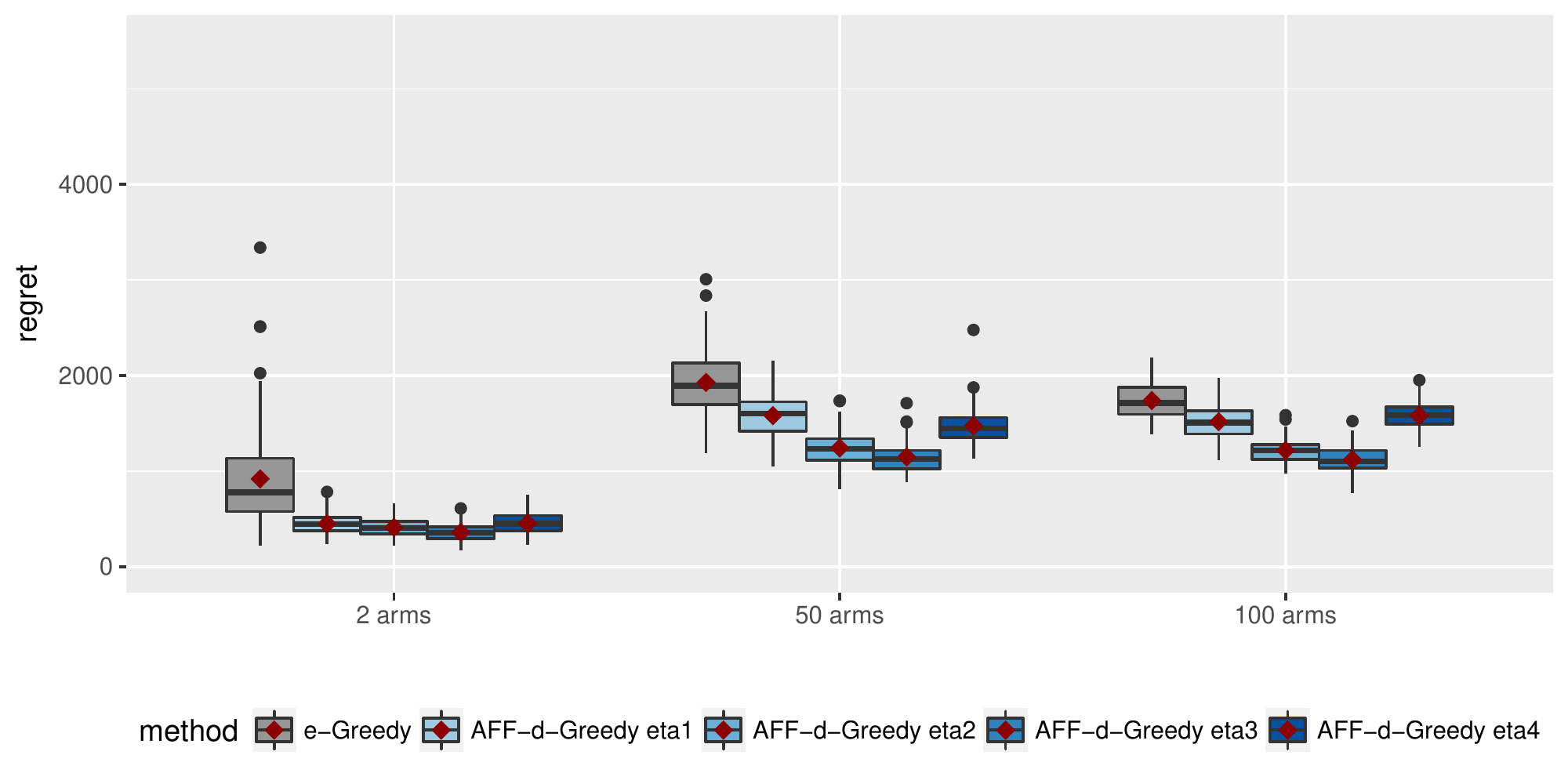}
    \caption{AFF-$d$-Greedy algorithm with different $\eta$ values. $\eta_{1} = 0.0001, \eta_{2} = 0.001$, $\eta_{3} = 0.01$, and $\eta_{4}(t) = 0.0001/s^{2}_{t}$, where $s^{2}_{t}$ is as in (\ref{eqt-AFFvariance}).}
    \label{plot-RWRB-multi-Data1-greedy} 
\end{figure}

\begin{figure}[H]
    \centering
    \subfloat[AFF-UCB1]
    {\includegraphics[width=0.9\textwidth, height=0.17\textheight]{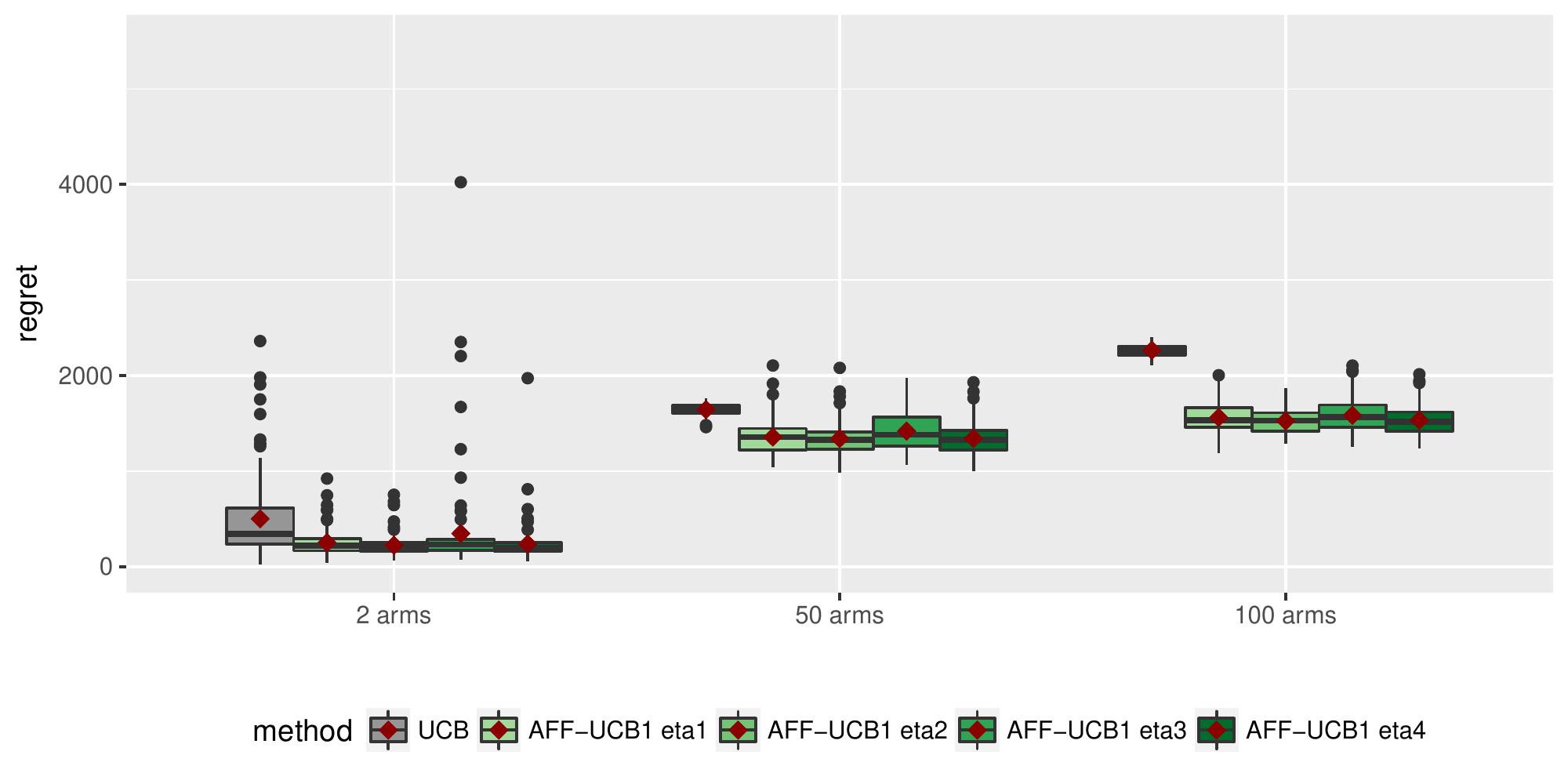}}\hfill
    \subfloat[AFF-UCB2]
    {\includegraphics[width=0.9\textwidth, height=0.17\textheight]{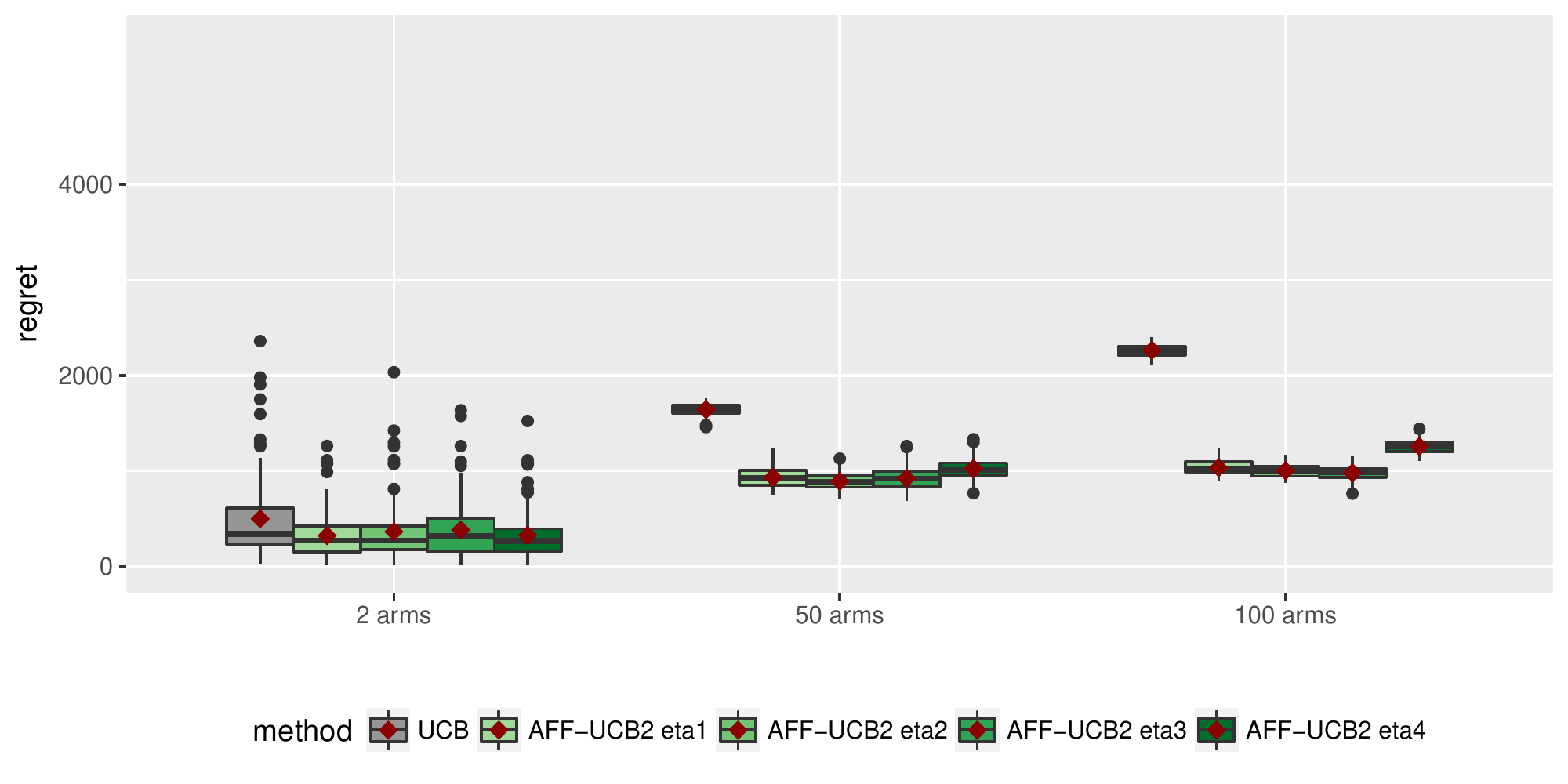}}\hfill
    
    \caption{AFF versions of UCB algorithm with different $\eta$ values. $\eta_{1} = 0.0001$, $\eta_{2} = 0.001$, $\eta_{3} = 0.01$, and $\eta_{4}(t) = 0.0001/s^{2}_{t}$, where $s^{2}_{t}$ is as in (\ref{eqt-AFFvariance}).}
    \label{plot-RWRB-multi-Data1-UCB} 
\end{figure}

\begin{figure}[H]
    \centering
    \subfloat[AFF-TS]
    {\includegraphics[width=0.9\textwidth, height=0.17\textheight]{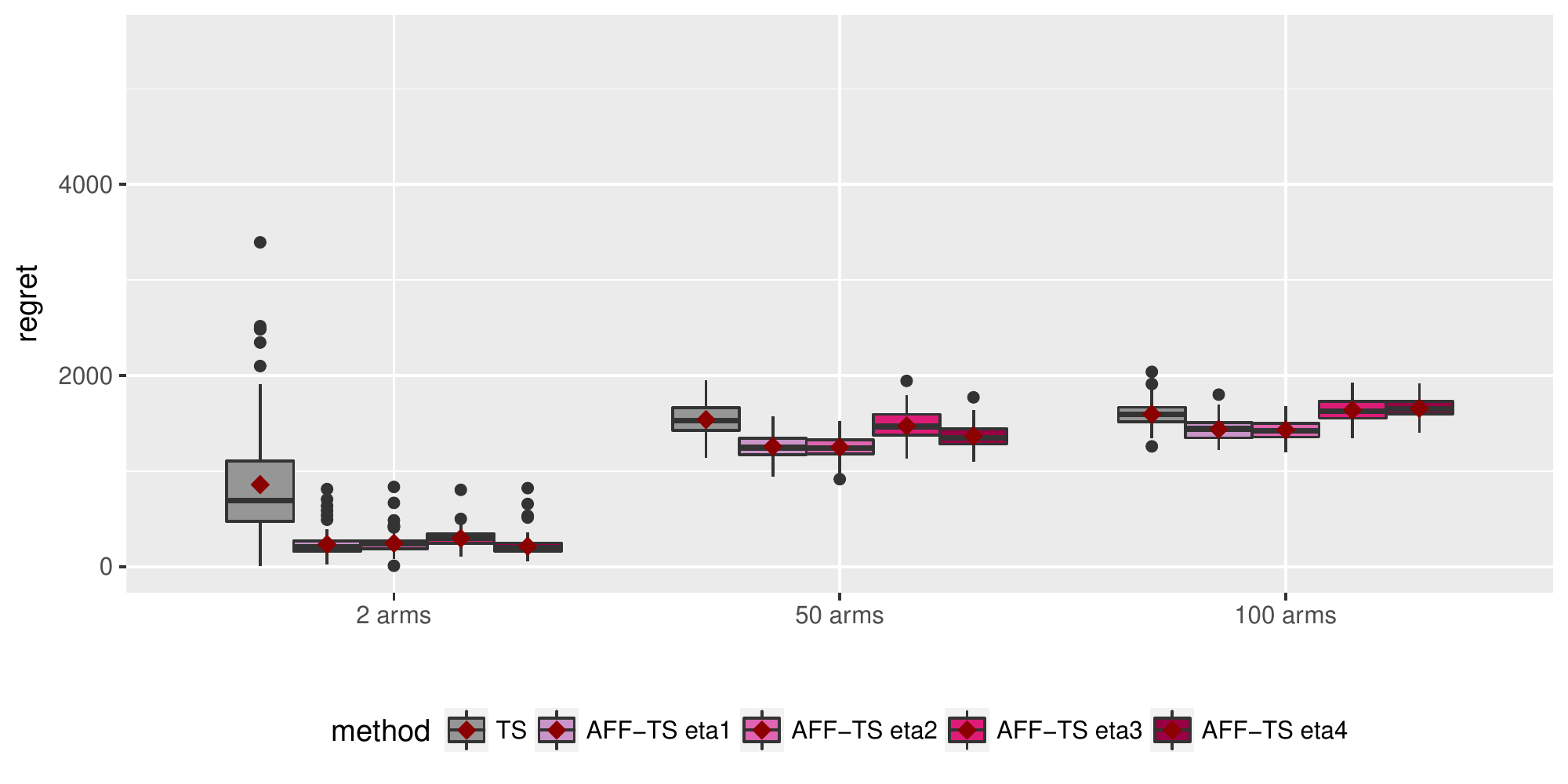}}\hfill
    \subfloat[AFF-OTS]
    {\includegraphics[width=0.9\textwidth, height=0.17\textheight]{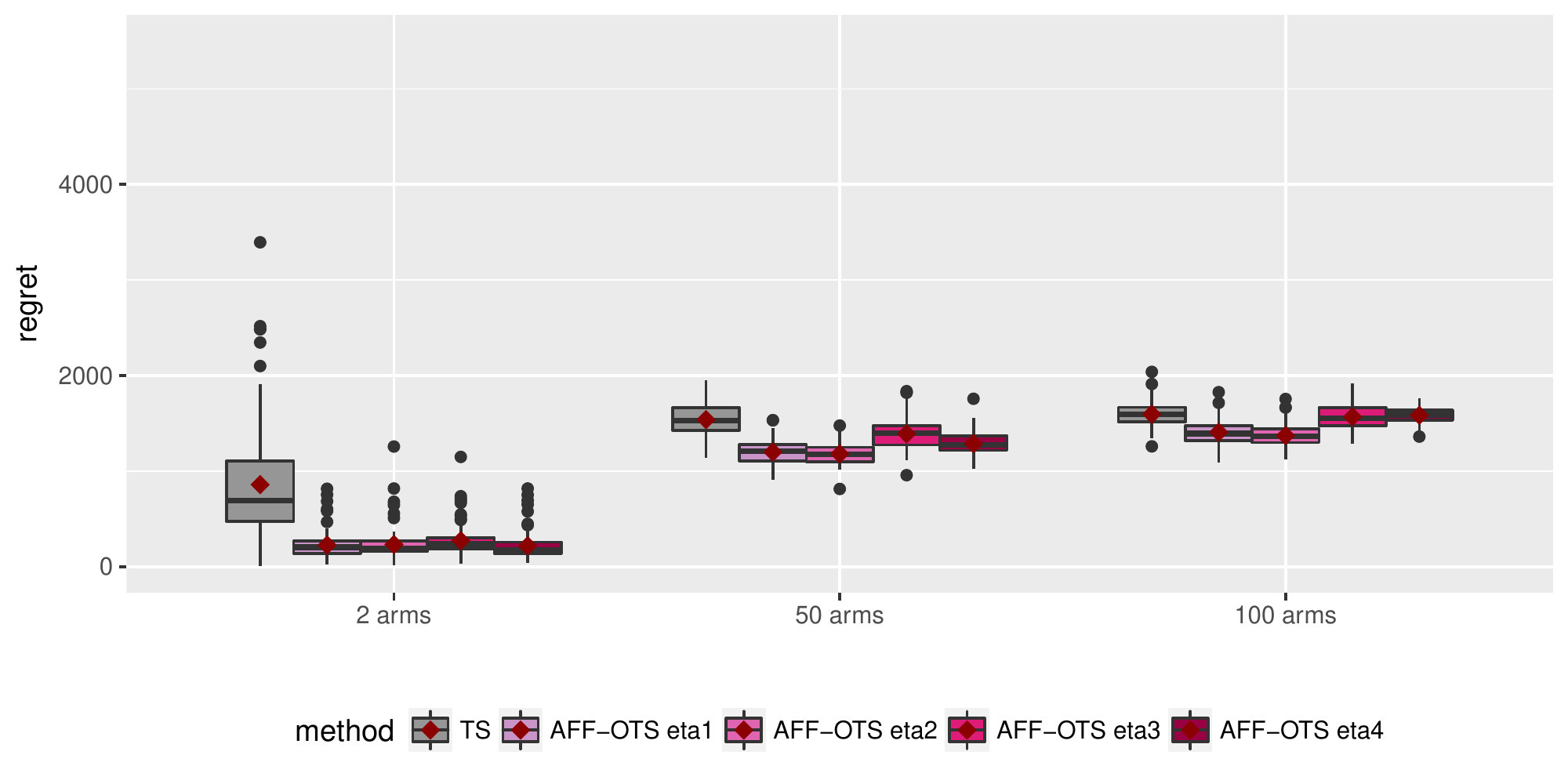}}\hfill
    
    \caption{AFF versions of TS algorithm with different $\eta$ values. $\eta_{1} = 0.0001$, $\eta_{2} = 0.001$, $\eta_{3} = 0.01$, and $\eta_{4}(t) = 0.0001/s^{2}_{t}$, where $s^{2}_{t}$ is as in (\ref{eqt-AFFvariance}).}
    \label{plot-RWRB-multi-Data1-TS} 
\end{figure}

\begin{figure}[H]
    \centering
    \subfloat[D-UCB: $\lambda_{1} = 1-(4)^{-1}\sqrt{ \Upsilon_{T}/T}$, $\lambda_{2} = 0.99$, $\lambda_{3} = 0.8$, and $\lambda_{4}=0.5$.]
    {\includegraphics[width=0.9\textwidth, height=0.17\textheight]{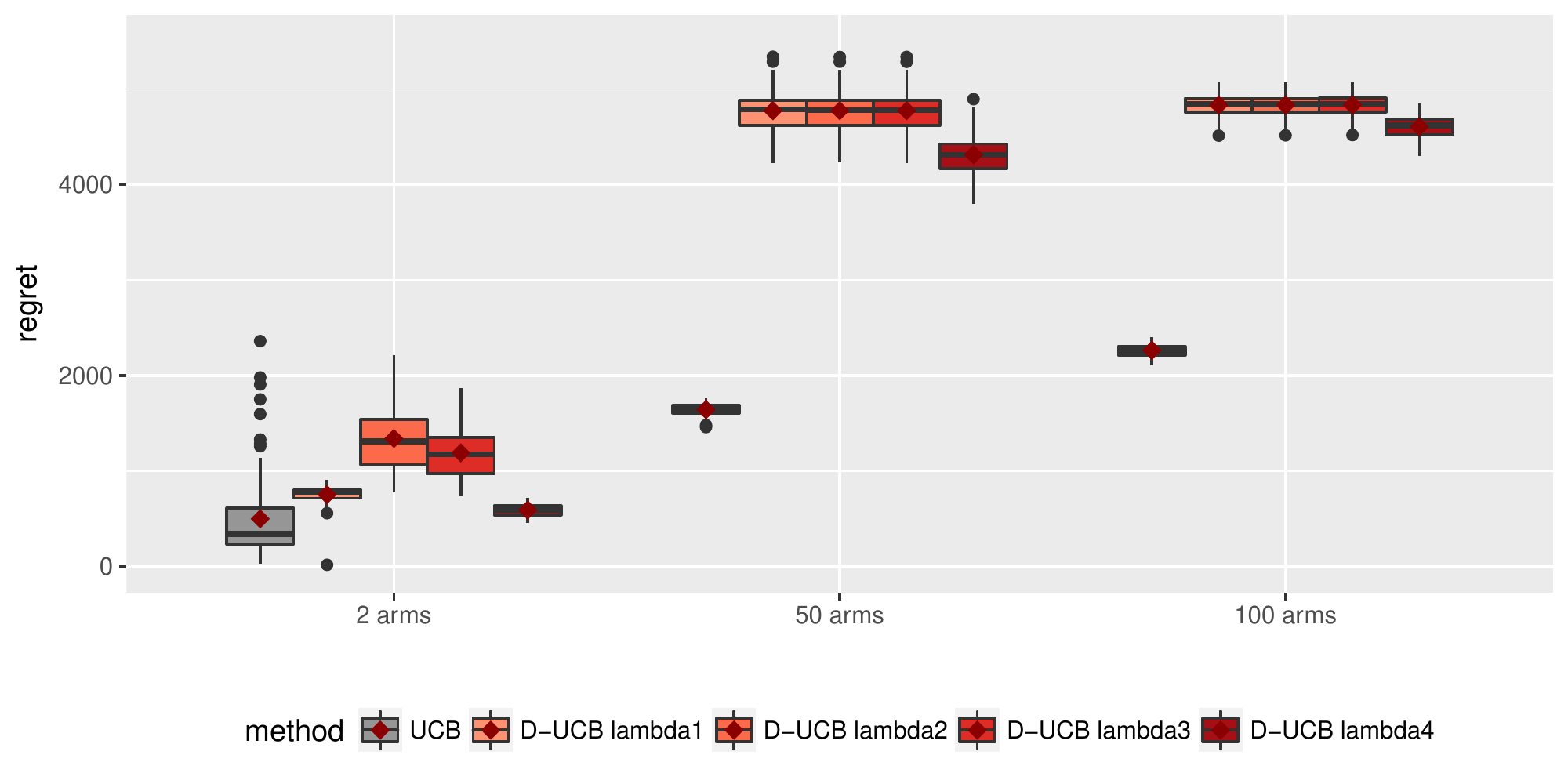}}\hfill
    \subfloat[SW-UCB: $W_{1} = 2
\sqrt{T\log(T )/ \Upsilon_{T}}$, $W_{2} = 10$, $W_{3} = 100$, and $W_{4}=1000$.]
    {\includegraphics[width=0.9\textwidth, height=0.17\textheight]{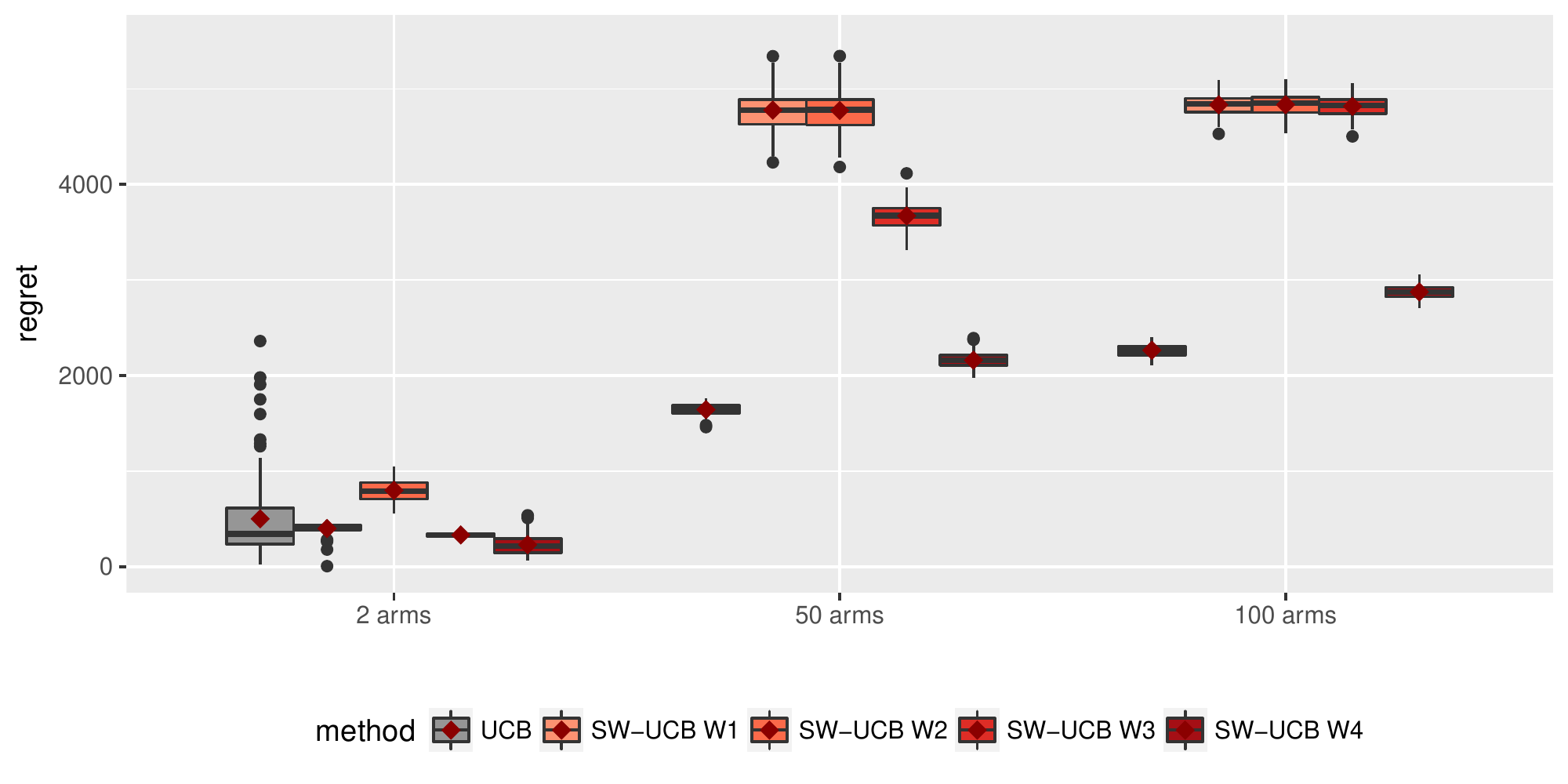}}\hfill
    
    \caption{D-UCB and SW-UCB algorithms with different values of key parameters.}
    \label{plot-RWRB-multi-Data1-DUCB} 
\end{figure}

In this section, we examine the influence of the step size $\eta$ on the AFF MAB algorithms. We present only for Case~3 (see Section \ref{sec-Drifting}) for the sake of brevity; results for other cases are very similar and hence omitted. For each AFF MAB algorithm, we do experiments with $\eta_{1} = 0.0001$, $\eta_{2} = 0.001$, $\eta_{3} = 0.01$, and $\eta_{4}(t) = 0.0001/s^{2}_{t}$, where $s^{2}_{t}$ is the AFF variance defined in (\ref{eqt-AFFvariance}). Note here $\eta_1$, $\eta_2$, and $\eta_3$ are fixed, while $\eta_{4}$ can change over time. In addition, we compare the influence of key parameters in D-UCB and SW-UCB. For D-UCB, we choose 4 values of the fixed discounting factor which are $\lambda_{1} = 1-(4)^{-1}\sqrt{ \Upsilon_{T}/T}$, $\lambda_{2} = 0.99$, $\lambda_{3} = 0.8$, and $\lambda_{4}=0.5$. For SW-UCB, 4 different window sizes are $W_{1} = 2
\sqrt{T\log(T )/ \Upsilon_{T}}$, $W_{2} = 10$, $W_{3} = 100$, and $W_{4}=1000$. $\Upsilon_{T}$ is the total number of switch points during the total $T$ rounds and its value can be different for individual replications.
Figures~\ref{plot-RWRB-multi-Data1-greedy}-\ref{plot-RWRB-multi-Data1-TS} display the results for AFF-$d$-Greedy, AFF-UCB1/AFF-UCB2, and AFF-TS/AFF-OTS respectively. We can see that the algorithms are not particularly sensitive to the step size $\eta$. The results for D-UCB and SW-UCB can be found in Figure~\ref{plot-RWRB-multi-Data1-DUCB}. It can be seen that, the results vary more for these two algorithms. We can conclude that it is easier to tune $\eta$ than the fixed discounting factor $\lambda$ or window size $W$.

%
%
%
%

\subsubsection{Using Adaptive Forgetting Factors to Tune Parameter $C$ in Dynamic Thompson Sampling}
\label{sec-exp-AFFDTS}

In Section \ref{sec-TuneC}, we discussed the use of adaptive estimation to tune the input parameter $C$ in the DTS algorithm proposed by \citet{Gupta2011}, and we offered two self-tuning solutions, AFF-DTS1 and AFF-DTS2. We use the two-armed abruptly changing example (Case~1 in Section \ref{sec-AbruptChange}) to illustrate how the AFF version algorithms can reduce the sensitivity to $C$.

\begin{figure}[H]
    \centering
         \includegraphics[width=0.7\textwidth, height=0.22\textheight]{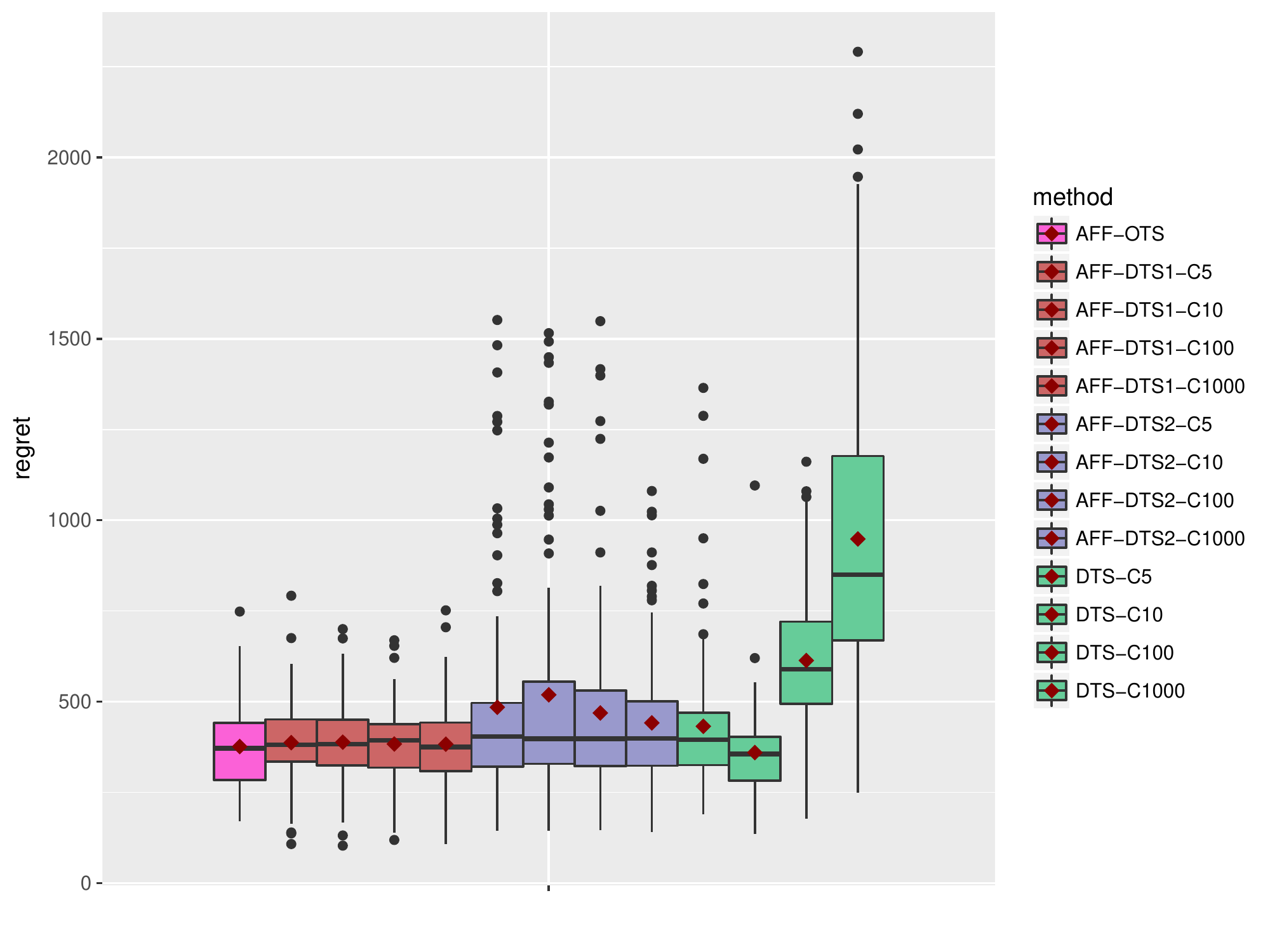}
    \caption{Boxplot of total regret for algorithms DTS, AFF-DTS1, and AFF-DTS2. Acronym like DTS-C5 represents the DTS algorithm with parameter $C=5$. Similarly acronym like AFF-DTS1-C5 represents the AFF-DTS1 algorithm with initial value $C_{0}=5$. The result of AFF-OTS is plotted as a benchmark.}
    \label{plot-DTS-Uniform} 
\end{figure}

We test $C = 5$, 10, 100, and 1000 for DTS, AFF-DTS1, and AFF-DTS2. It (the $C$ value) works as the initial value of $C_{t}$ in AFF-DTS1 and AFF-DTS2. Step size $\eta=0.001$ is used for AFF related algorithms.
Figure~\ref{plot-DTS-Uniform} displays the boxplot of total regret. We also plotted the result of AFF-OTS as a benchmark since it has good performance in all cases studied in the previous section. 
From Figure~\ref{plot-DTS-Uniform}, the performance of AFF-DTS1 and AFF-DTS2 are very stable, while DTS is very sensitive to $C$. With a bad choice of $C$ (i.e., 100 and 1000 in this case), the total regret of DTS is much higher than AFF-DTS1 and AFF-DTS2.

\section{Conclusion}
\label{sec-con}
We have seen that the performance of popular MAB algorithms can be improved significantly using AFFs.  The improvements are substantial when the arms are not distinguishable in the long run, i.e., the arms have the same long-term averaged expected reward $\bar{\mu}_{T}$, $\bar{\mu}_{T} = \mathbb{E}[\frac{1}{T} \sum_{i=1}^{T} \mu_{i}]$. For the case that one arm has a higher $\bar{\mu}_{T}$ (e.g., the two-armed example in Case~2), gains for the AFF MAB algorithms seem marginal, but there is no loss in performance, so practitioners could be encouraged to implement our adaptive methods when they do not have knowledge of the behaviour of $\mu_t$ with time. In addition, the performance gains for a large number of arms are very pronounced for all AFF MAB methods, and the performance is much better than SW-UCB and D-UCB. Finally, the AFF MAB algorithms we proposed are easy to implement; they do not require any prior knowledge about the dynamic environment, and seem to be more robust to tuning parameters.


Our methods AFF-UCB1/AFF-UCB2 and AFF-TS are similar to others that deployed a fixed forgetting factor: D-UCB \citep{Garivier2011} and discounted Thompson sampling \citep{Raj2017} respectively. However, using an adaptive forgetting factor is more flexible as fixed forgetting factors correspond to the assumption of a fixed speed that $\mu_{t}$ evolves (drifting cases) whereas adaptive forgetting factor can also handle the case of varying speed of drift (e.g., abrupt change is a special case), wherein the optimal forgetting factor is not only unknown but itself time-varying.
Combining adaptive estimation with UCB was more challenging. The reason was that one needs to reinterpret the estimate of $\mu_{t}$ from a stable long run average to a ``more dynamic'' estimator (with less memory), and modify accordingly the upper bound. To boost the exploration of unobserved arms, we use the extra information from AFF updating and modified the decision bound/posterior distribution. Our AFF-deployed UCB and TS algorithms increase the probability of selecting an idle arm according to its time of being idle, and decrease that probability with the total number of arms. This exploration scheme works very well especially in the numerical study of high number of arms. For the UCB method, we provide two AFF versions: AFF-UCB1 and AFF-UCB2. They both work well (and AFF-UCB2 is better) when the number of arms is large while AFF-UCB1 is more stable in cases of small number of arms. Unlike the UCB and TS cases, our AFF version of $\epsilon$-Greedy method do not deploy a particular exploration boosting scheme for unselected arms in decision. The main reason is that, in the original $\epsilon$-Greedy algorithm, it does not involve the uncertainty in decision, and there is no obvious way to include that for this type of method. However, our AFF-$d$-Greedy method still improves the performance in most cases.


We conclude by mentioning some interesting avenues for future work. One extension is to apply AFF-based methods for more challenging problems, e.g., rotting bandits \citep{Levine2017}, contextual bandits \citep{Langford2008, Li2010}, and applications like online advertising. Another extension could involve a rigorous analysis of how the bias in AFF estimation varies with time and how can this affect the selection in MAB problems.

\newpage

\appendix
\section{Deriving the Exploration Bonus that Corresponds to the AFF Mean $\hat{Y}_{t}$}
\label{app-AFFUCB}


We present how we derive the corresponding exploration bonus $B_{t}$,\begin{align*}
B_{t} = \sqrt{\frac{- \log (0.05)}{2 (w_{t})^2/ k_{t} }},
\end{align*}
for the AFF mean $\hat{Y}_{t}$ defined in (\ref{eqt-AFFmean}).

According to (\ref{eqt-AFFmean})-(\ref{eqt-wt}), the AFF mean $\hat{Y}_{t}$ for the independent reward stream $Y_{1}, \cdots, Y_{t}$ is: 
\begin{align*}
\hat{Y}_{t}  =  \frac{m_{t}}{w_{t}},
~~~~ m_{t}  = \sum_{i=1}^{t} \left(\prod_{p=i}^{t-1} \lambda_{p} \right) Y_{i},
~~~~ w_{t} = \sum_{i=1}^{t} \left(\prod_{p=i}^{t-1} \lambda_{p} \right).
\end{align*} 

According to Hoeffding's inequality, we have
\begin{align}
\notag
P(\text{E}[\hat{Y}_{t}] - \hat{Y}_{t} \geq B_{t}) 
&= P \left(\text{E} \left[ \frac{m_{t}}{w_{t}} \right] - \frac{m_{t}}{w_{t}} \geq B_{t} \right)\\
\notag
& = P \left( \text{E}[m_{t}] - m_{t} \geq B_{t} w_{t} \right) \\
\notag
& \leq \exp \left(-\frac{2(B_{t})^{2} (w_{t})^2}{k_{t}} \right).
 \end{align}
 where $k_{t} = \sum_{i=1}^{t} (\prod_{p=i}^{t-1} \lambda_{p}^{2})$.
We use the fact that $\left(\prod_{p=i}^{t-1} \lambda_{p} \right) Y_{i}$ is bounded in $\left [0, \left(\prod_{p=i}^{t-1} \lambda_{p} \right) \right ]$ since $Y_{i}$ is bounded in $[0,1]$. 
Whilst the use of Hoeffding's inequality is typically for i.i.d variables, there are similar expressions for Markov chains \citep{Glynn2002}, which fits to our framework.

Let $\xi$ denote $P(\text{E}[\hat{Y}_{t}] - \hat{Y}_{t} \geq B_{t})$, and set $\xi = \exp \left(-\frac{2(B_{t})^{2} (w_{t})^2}{k_{t}} \right)$, we have 
$$B_{t} = \sqrt{-\frac{\log(\xi) k_{t}}{2 (w_{t})^2}}. $$
$\xi$ is the the probability that the difference between $\text{E}[\hat{Y}_{t}]$ and  $\hat{Y}_{t}$ exceeds $B_{t}$. The form of $B_{t}$ is similar to the exploration bonus in UCB \citep{Auer2002c}, and in UCB, $\xi$ was set to $\xi=t^{-4}$ to obtain a tighter upper bound as the number of trials increases (that is, exploration is reduced over time). This is sensible in static cases as the estimates converge with $t$. However, as we are interested in dynamic cases, we are in favour of a bound that keeps a certain level of exploration over time, and hence we take a constant $\xi=0.05$, and get the exploration bonus 
$$B_{t} = \sqrt{\frac{- \log (0.05)}{2 (w_{t})^2/ k_{t} }}.$$

\newpage
\bibliography{AFFMAB}
\end{document}